\documentclass[runningheads]{llncs}
\usepackage{graphicx}

\usepackage{enumitem}

\usepackage{epsfig}
\usepackage{graphicx}
\usepackage{float}
\usepackage{wrapfig}
\usepackage{amsmath,amssymb}
\usepackage{algorithm,algorithmicx,algpseudocode}
\usepackage{authblk}
\usepackage{bm,xspace}
\usepackage{comment}
\usepackage{verbatim}
\usepackage{multirow}
\usepackage{balance}
\usepackage{url}
\usepackage{booktabs}
\usepackage{etoolbox,siunitx}
\usepackage{calc}
\usepackage{pifont,hologo}
\usepackage{nicefrac}
\usepackage{wrapfig}
\usepackage[symbol]{footmisc}

\makeatletter
\let\NAT@parse\undefined
\makeatother

\usepackage[super]{nth}
\usepackage{nicefrac}
\robustify\bfseries

\def\nn{\mathbf{n}}

\def\pp{\mathbf{p}}
\def\qq{\mathbf{q}}

\def\tt{\mathbf{t}}

\def\xx{\mathbf{x}}
\def\yy{\mathbf{y}}

\def\PP{\mathbf{P}}
\def\QQ{\mathbf{Q}}
\def\RR{\mathbf{R}}

\def\cC{\mathcal{C}}

\def\lL{\mathcal{L}}

\DeclareMathOperator*{\argmin}{arg\,min}

\newcommand\llesser{\mathbin{<\!\!\!<}}

\DeclareMathSymbol{@}{\mathord}{letters}{"3B}

\def\latex/{\LaTeX}
\def\bibtex/{\hologo{BibTeX}}

\newcommand\thetalut{\bm{\theta}_{\text{lut}}}
\newcommand\philut{\bm{\phi}_{\text{lut}}}

\newcommand{\eg}{\textit{e}.\textit{g}.}

\newdimen\parboxheight

\makeatletter
\newcommand*{\org@iiiparbox}{}
\let\org@iiiparbox\@iiiparbox
\renewcommand*{\@iiiparbox}[2]{%
  \ifx\relax#2%
    \setlength{\parboxheight}{0pt}%
  \else
    \setlength{\parboxheight}{#2}%
  \fi
  \org@iiiparbox{#1}{#2}%
}
\makeatother

\newcommand\rotentry[2]{\parbox[t]{2mm}{\rotatebox[origin=l]{#1}{#2}}}

\newcommand\ssbf[1]{\scriptsize{\textbf{#1}}}
\newcommand\ssf[1]{\scriptsize{#1}}

\usepackage[pagebackref=true,breaklinks=true,letterpaper=true,colorlinks=false,bookmarks=false]{hyperref}
\usepackage{cleveref}[2012/02/15]
\usepackage{balance}
\usepackage{capt-of}
\usepackage{cite}
\usepackage{tikz}
\usepackage{comment}
\usepackage{amsmath,amssymb} 
\usepackage{color}

\usepackage[accsupp]{axessibility}  

\begin{document}
\pagestyle{headings}
\mainmatter
\def\ECCVSubNumber{1488}  

\title{Revisiting LiDAR Registration and Reconstruction: A Range Image Perspective} 

\renewcommand*{\thefootnote}{\fnsymbol{footnote}}

\titlerunning{Revisiting LiDAR Registration and Reconstruction}
%
\author{Wei Dong  \inst{1,}\footnote[1]{indicates equal contribution.} \and
Kwonyoung Ryu\inst{2,*} \and
Michel Kaess\inst{1} \and
Jaesik Park\inst{2}}
\authorrunning{W. Dong et al.}
%
\institute{\footnotesize Carnegie Mellon University, Pittsburgh, PA, USA. \and
POSTECH, Pohang, South Korea.}
\maketitle
\vspace{-7mm}
\begin{abstract}
Spinning LiDAR data are prevalent for 3D vision tasks. 
Since LiDAR data is presented in the form of point clouds, expensive 3D operations are usually required. This paper revisits spinning LiDAR scan formation and presents a cylindrical range image representation with a ray-wise projection/unprojection model. It is built upon raw scans and supports lossless conversion from 2D to 3D, allowing fast 2D operations, including 2D index-based neighbor search and downsampling
We then propose, to the best of our knowledge, the first multi-scale registration and dense signed distance function (SDF) reconstruction system for LiDAR range images. We further collect a dataset of indoor and outdoor LiDAR scenes in the posed range image format. A comprehensive evaluation of registration and reconstruction is conducted on the proposed dataset and the KITTI dataset. Experiments demonstrate that our approach outperforms surface reconstruction baselines and achieves similar performance to state-of-the-art LiDAR registration methods, including a modern learning-based registration approach. Thanks to the simplicity, our registration runs at 100Hz and SDF reconstruction in real time. The dataset and a modularized C++/Python toolbox will be released.

%

\vspace{-1mm}
\vspace{-3mm}
\end{abstract}

\renewcommand*{\thefootnote}{\arabic{footnote}}
\setcounter{footnote}{0} 

\section{Introduction}
\label{sec:introduction}
LiDAR scanners are prevalent sensors used to obtain range data and provide 3D geometry by measuring the time of flight of modulated laser pulses. Compared with camera-like solid-state LiDARs, spinning LiDARs capture full $360^\circ$ views, thus they are widely applicable to robotics, remote sensing, and autonomous driving. 
Popular spinning LiDARs such as Velodyne~\cite{velodyne2007} and Ouster~\cite{ouster2018} are designed in a similar fashion: a line of scan is measured vertically; a complete $360^\circ$ scan is formed by horizontally spinning the sensor to accumulate line scans in a consistent coordinate system.

\begin{figure}[t]
    \centering
    \includegraphics[width=.98\columnwidth]{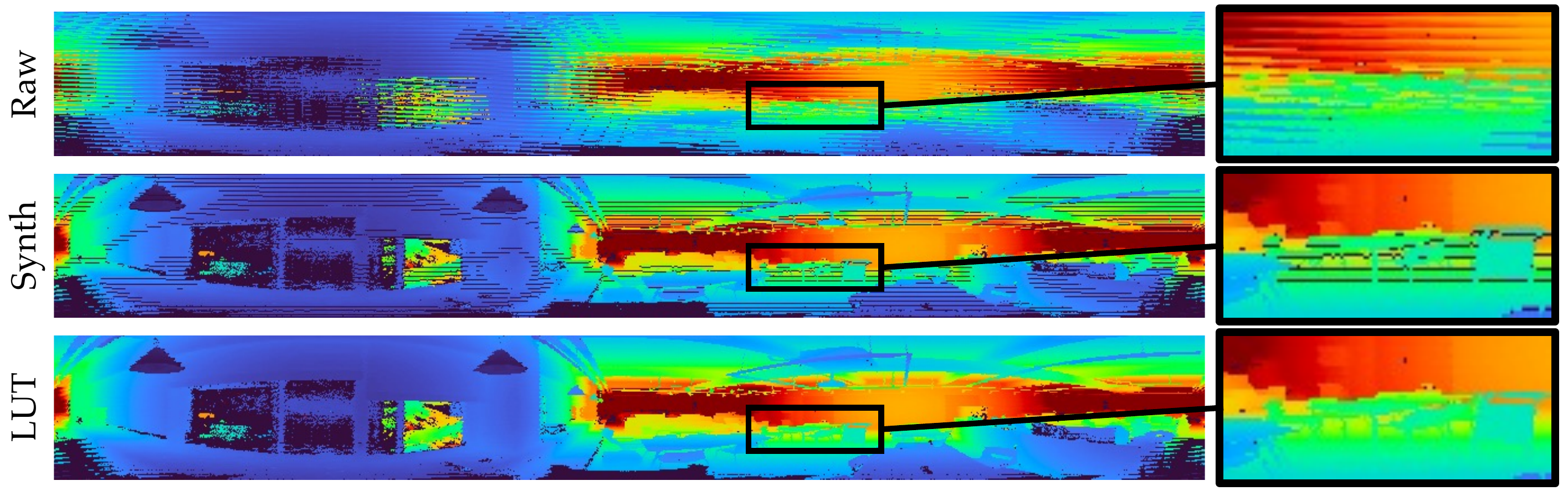}
    \caption{Visualization of a LiDAR scan as a cylindrical range image in various forms. {Synthetically projecting} a point cloud to a cylindrical image~\cite{Behley2018EfficientSS,shan2018lego} results in artifacts (middle) due to inaccurate altitude mapping. The cylindrical image view (bottom) of {raw scans} (top) with a lookup table (LUT) is loseless. }
    \vspace*{-3mm}
    \label{fig:formulation}
\end{figure}

It is clear that intrinsic geometric transformations exist in the conversion from raw scans to a 3D point cloud, consisting of spherical projective and rigid transformations. Yet the value of low-level conversions are down-weighted for convenience, and many hardware drivers and downstream datasets~\cite{Geiger2012CVPR, tan2020toronto3d, hackel2017semantic3d} only provide \emph{3D point clouds} to the user. 
There is an advantage of the design, since that prevalent 3D data format is acceptable to most 3D processing pipelines. However, the intrinsic relations between the scanned points are discarded, and k-d trees~\cite{bentley1975kdtree} have to be constructed to find nearest neighbors in the Euclidean 3D space, which require highly optimized implementation for real-time systems such as LiDAR odometry (LO) and simultaneous localization and mapping (SLAM).
Recent studies~\cite{Behley2018EfficientSS, shan2018lego} generate proxy 2D range images from point clouds via synthetic projections to accelerate neighbor search, reducing the query complexity from $O(N \log N)$ to $O(N)$ for a point cloud of size $N$ with the drop of a k-d tree. Yet further advantages of the image representation, from fast down-sampling to signed distance computation, are not well-studied; a loss of data quality is also inevitable due to synthetic projection, as shown in Fig.~\ref{fig:formulation}.

A cylindrical image view of spinning LiDAR's \emph{raw scan lines}, on the other hand, is efficient without losing the data quality against its geometry-equivalent point cloud. By nature, it supports fast projective data association~\cite{whelan2015elasticfusion} and neighbor search in images. Therefore, direct visual odometry~\cite{whelan2015elasticfusion,newcombe2011kinectfusion} and signed distance function (SDF) reconstruction~\cite{curless1996volumetric, newcombe2011kinectfusion} are applicable. 

In this paper, we revisit the LiDAR\footnote{Without confusion, we regard camera-like solid-state LiDARs as Depth sensors in contrast to spinning LiDARs. The term \emph{LiDAR} specifically denotes spinning LiDARs in the rest of the paper.} data formulation, and propose the range image-based representation shown in Fig.~\ref{fig:formulation}. 
Our contributions can be summarized as:
\renewcommand{\labelitemi}{$\bullet$}
\begin{itemize}[leftmargin=*]
    \item A cylindrical image view of LiDAR data directly from the scan lines along with an intrinsic spherical projective model that supports accurate conversions between 2D and 3D;
    \item Fast and effective multi-scale registration and scalable SDF reconstruction for range LiDAR images, accelerated on GPU. These operations are backward compatible to synthesized range images from point clouds, \eg, KITTI~\cite{Geiger2012CVPR};
    \item A new collection of LiDAR range image sequences of both indoor and outdoor scenes with pseudo ground truth poses, along with comprehensive evaluations on the task of registration and surface reconstruction.
\end{itemize}

\section{Related Work}
\label{sec:related}
\begin{figure*}[t]
\centering
\begin{tabular}{c}
  \includegraphics[width=\textwidth]{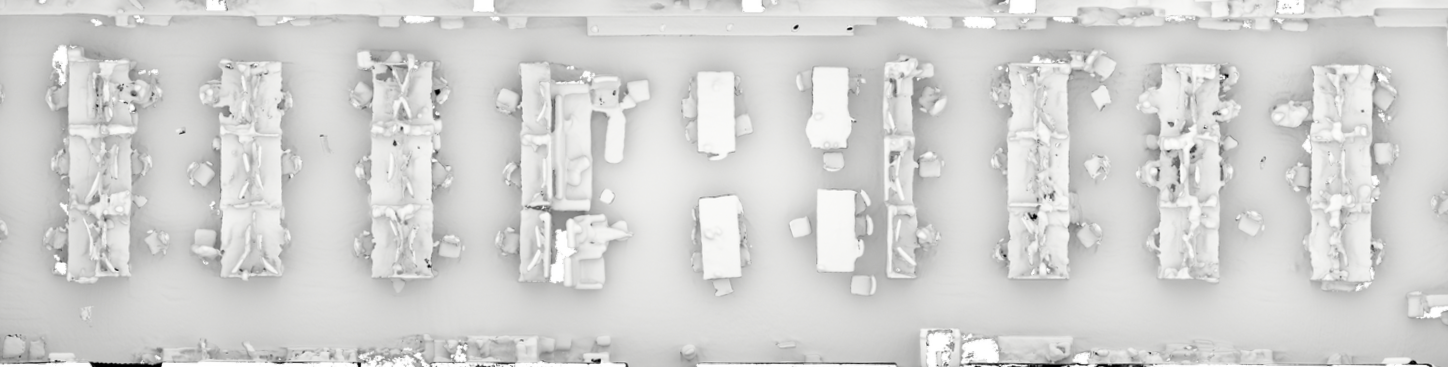} \\
\end{tabular}
\begin{tabular}{ccc}
  \includegraphics[height=1.4cm]{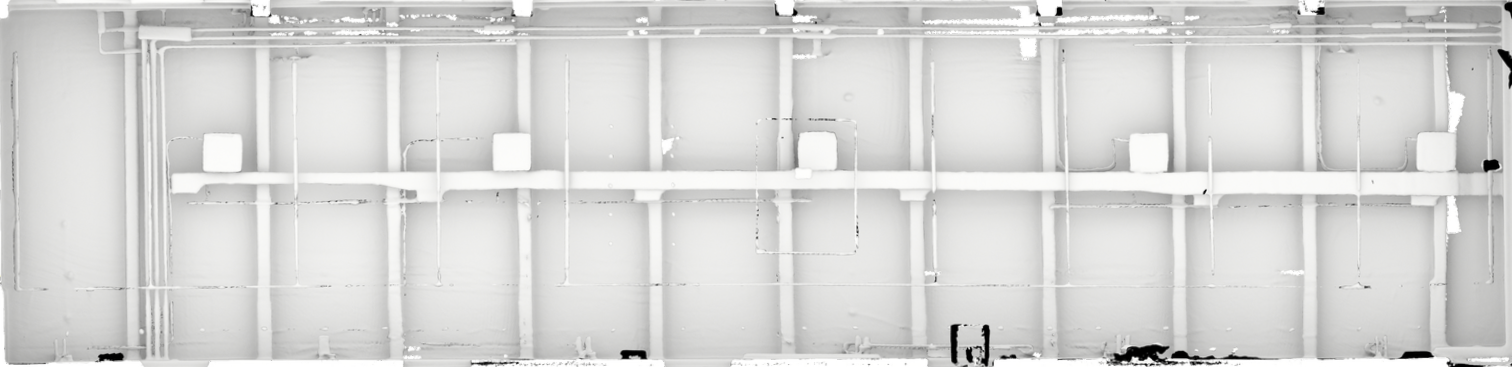} & \includegraphics[height=1.4cm]{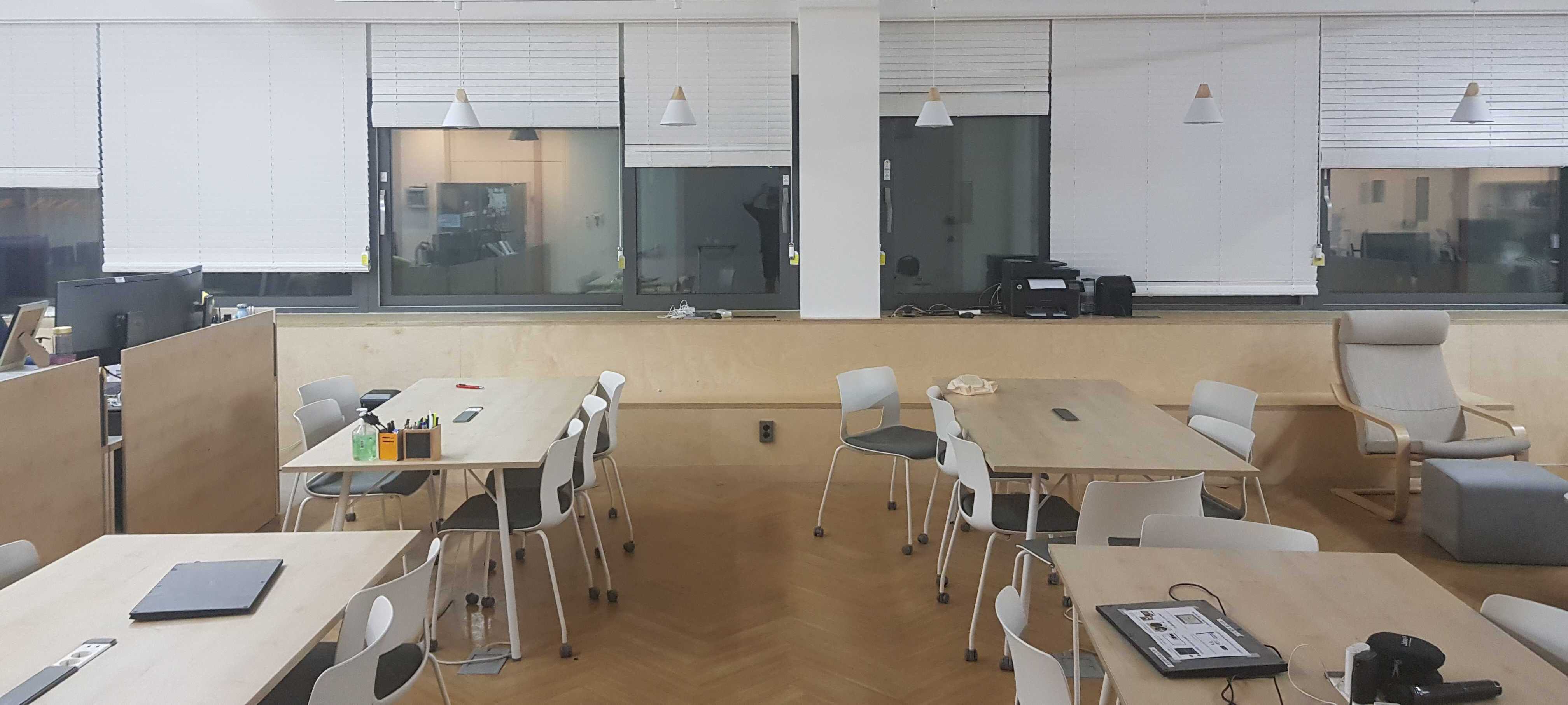} &
  \includegraphics[height=1.4cm]{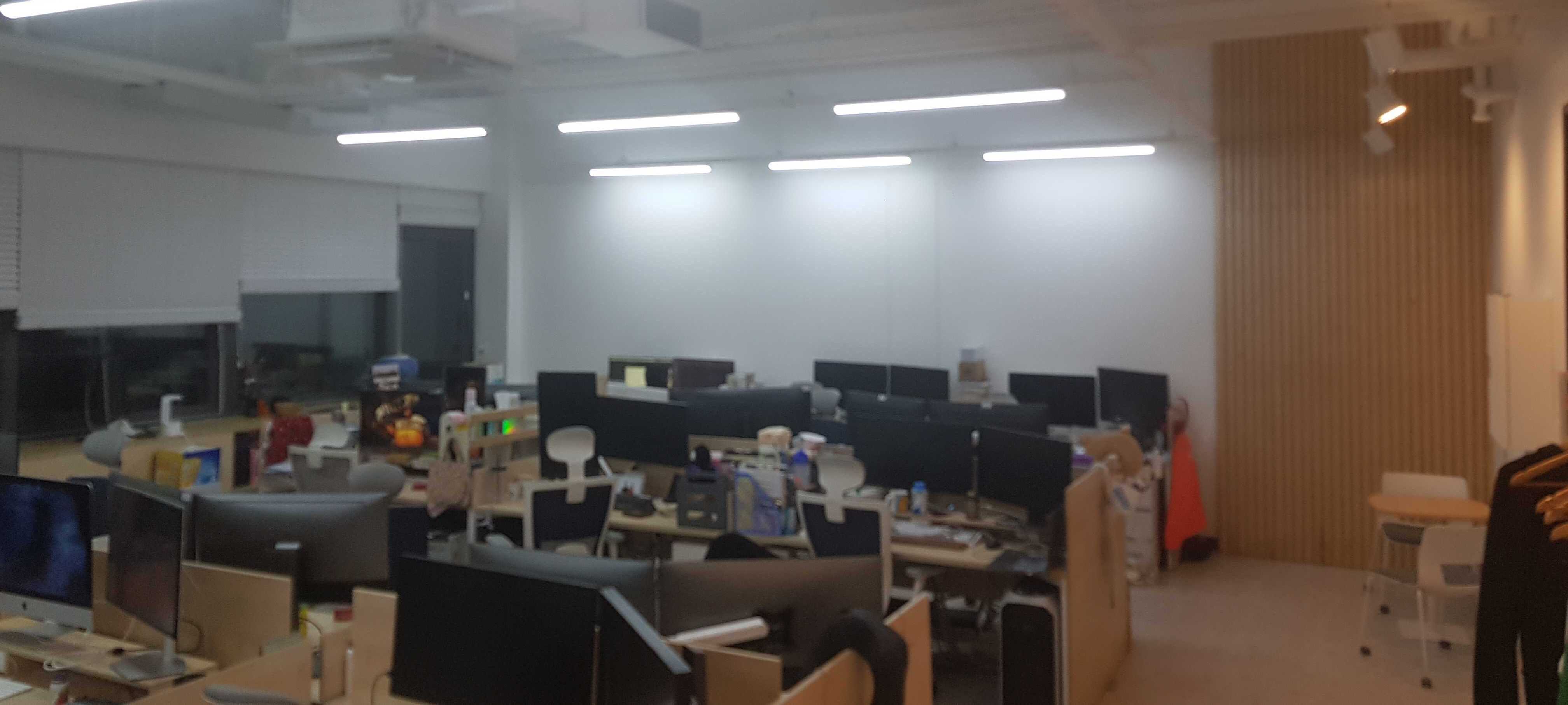}
\end{tabular}
\caption{Surface reconstruction via SDF integration of the \emph{lab} sequence from LiDAR range images. Top and bottom left are the ground and the ceiling rendered with Mitsuba 2~\cite{nimier2019mitsuba}. Bottom right are pictures of the scene.}
\vspace{-3mm}
\label{fig:teaser}
\end{figure*}

\noindent \textbf{Representation.} 
LiDAR data are generally viewed as point clouds, an unordered set of 3D points, the major format in prevalent LiDAR datasets~\cite{Geiger2012CVPR,tan2020toronto3d,hackel2017semantic3d}. Several datasets and systems~\cite{Geiger2012CVPR,bogoslavskyi2016fast,Behley2018EfficientSS,shan2018lego,chen2021range,Argoverse,nuScenes} project point clouds to the cylindrical image space and synthesize range images, but the data distribution is sparse with significant artifacts.
In RGB-D cameras, however, 3D point clouds are densely packed as 2D images~\cite{sturm12iros, Choi2015, park2017colored, dai2017scannet}, also known as organized point clouds~\cite{rusu20113d}. The major difference comes from the hardware. LiDARs rely on the rotation of a line scanner, hence the output is more likely to be interpreted as an unordered set, whereas RGB-D scanners use structured sensors that by nature capture images. 
An image formation allows efficient operations by indexing with coordinates, while a point cloud requires trees~\cite{meagher1982octree,bentley1975kdtree} or spatially hashed voxels~\cite{niessner2013real} to enable fast accessing by location.

\noindent \textbf{Registration.}
Point cloud registration is a well-studied topic that aligns two point sets with a known initial pose. In the point cloud format, variations of iterative closest points (ICP) are classical solutions~\cite{bes1992,rusinkiewicz2001,segal2009generalized, park2017colored}. These methods depend on nearest neighbor search in 3D using trees~\cite{bentley1975kdtree}, which is the bottleneck of the performance.
Learning-based algorithms~\cite{Wang_2019_ICCV,choy2020deep,bai2020d3feat,Choy_2020_CVPR, Lee2021DeepHV} seek to avoid nearest neighbor search via deep feature matching and/or the weighted Procrustes solver, but in practice require even more computation resources.
In the range image form, projective nearest neighbor is used instead to circumvent the 3D nearest neighbor search~\cite{newcombe2011kinectfusion,kerl13iros,whelan2015elasticfusion}. This formulation is introduced to LiDAR data~\cite{Behley2018EfficientSS, shan2018lego} by synthetically projecting point clouds to cylindrical images.

\noindent \textbf{Surface reconstruction.}
Conventional LiDAR reconstruction uses occupancy grids~\cite{hess2016real,hornung2013octomap}, where the space is coarsely divided into grids recording the occupancy probability. While it preserves the coarse 3D geometry, a dense surface reconstruction is often not applicable.
Several surfel based dense reconstruction algorithms exist for LiDARs~\cite{park2018elastic,Behley2018EfficientSS}, but they are hardware or system dependent and cannot be easily generalized; time-consuming triangulation is required to generate a mesh. Truncated SDF (TSDF) reconstruction~\cite{oleynikova2017voxblox,millane2018cblox, roldao20193d} has been adapted to LiDARs, but still relies on the point cloud representation with point-ray tracing, thus an adaptation to GPU is non-trivial due to the race conditions at ray intersections.
Surface reconstruction using depth images for RGB-D sensors is more flexible due to the calibrated pinhole camera model. In addition to surfel-based reconstruction~\cite{keller2013real,whelan2015elasticfusion}, dense volumetric TSDF reconstruction produces water-tight surfaces for medium to large scale scenes~\cite{newcombe2011kinectfusion, niessner2013real, Choi2015, park2017colored, dong2019gpu} and can function alone given pose and depth image inputs.

In this paper we represent \emph{raw scans} of LiDARs as cylindrical range images along with projective LiDAR intrinsics. We then propose efficient approaches for LiDAR range image based registration and reconstruction, accelerated on GPU. Due to the simplicity of the formulation, while retaining a similar accuracy, our approach is 15--50$\times$ faster in surface reconstruction, and 5--150$\times$ faster in registration.

\section{LiDAR Scans as Cylindrical Range Images}
\label{sec:formulation}
\begin{wrapfigure}{l}{.45\columnwidth}
  \vspace{-8mm}
    \centering
    \includegraphics[width=.45\columnwidth]{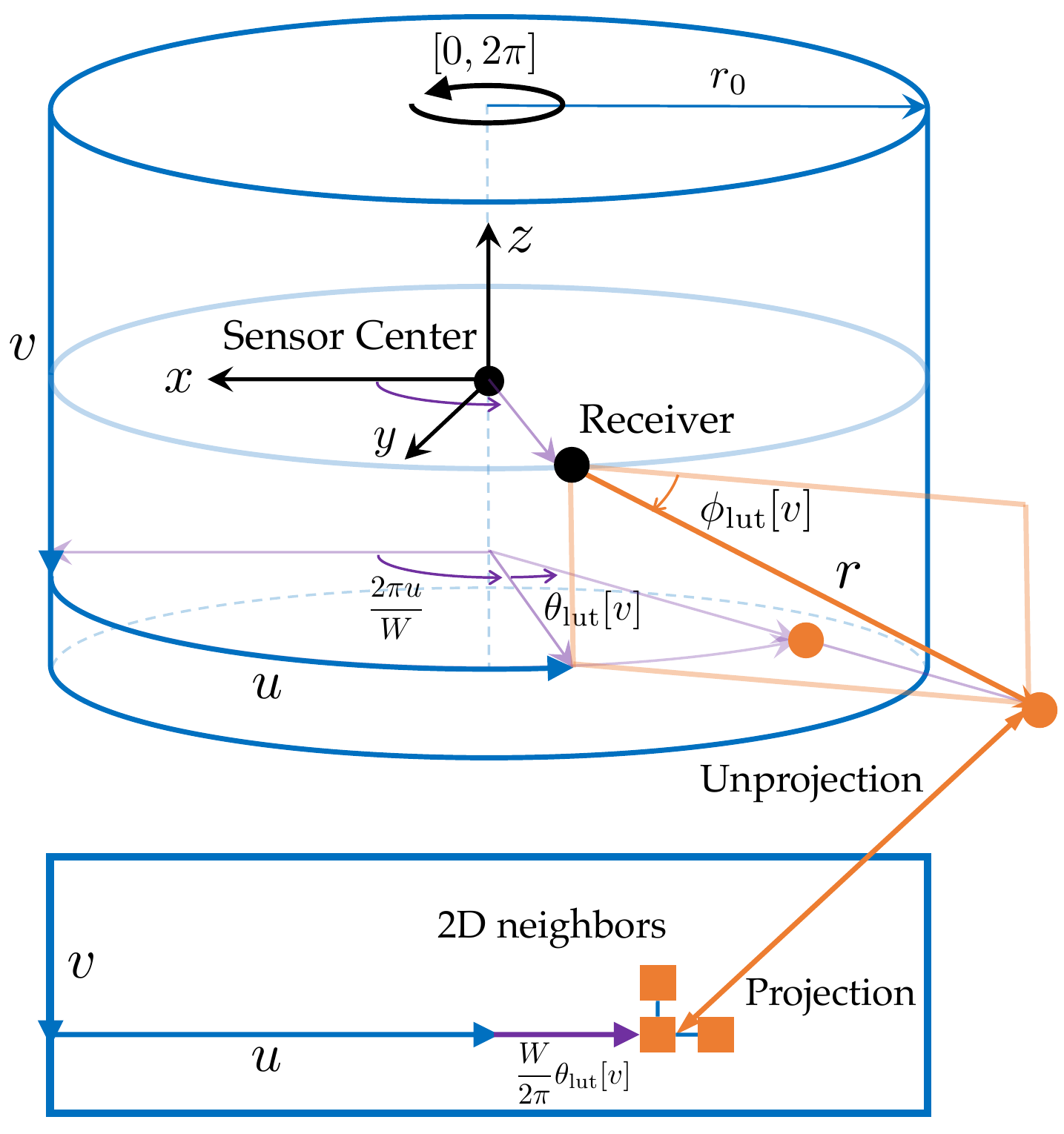}
    \caption{
    Illustration of the projection/unprojection procedure of a spinning LiDAR. Using the spherical projection model given the sensor intrinsics, 3D operations can be constrained on 2D range images with the routine of image processing.
    }
    \label{fig:illu}
    \vspace{-3mm}
\end{wrapfigure}

LiDAR scanners of our interest complete scans by rotation. A fixed number ($H$) of points are scanned roughly in a vertical line (corresponding to elevations) through aligned laser rays, and an accumulation of $W$ such lines form a complete scan spanning horizontally from $0^\circ$ to $360^\circ$. Therefore, a $H \times W$ range image can be naturally formed, where each pixel stores a range scalar associated to a ray. 

However, direct use of the raw LiDAR range map is not desirable. As shown in Fig.~\ref{fig:formulation}, the interlacing artifacts occur due to the local ray offset of each scan line. Hence, we need to adopt an azimuth intrinsic look-up table (LUT) $\thetalut$ provided by the manufacturer to compensate the offset. Similarly, a nonlinear elevation distribution associated with rays is defined by hardware design, in the form of another LUT $\philut$.
Given the ray-range image representation and the LUTs, we now analyze the spherical projection $\Pi: \mathbb{R}^3 \to S(3)$ and unprojection $\Pi^{-1}: \mathbb{R}^2 \times \Omega \to \mathbb{R}^3$ functions defined on the range image $\Omega$. We use $(u, v)$ to indicate a pixel coordinate, $r = \Omega(u, v)$ for the range reading, and $(x, y, z)$ for the corresponding 3D coordinate.

\subsection{Unprojection $\Pi^{-1}$}
A spinning LiDAR's receiver in charge of range sensing is located on a cylinder of radius $r_0$ enclosing the sensor center, see Fig.~\ref{fig:illu}. Therefore, the unprojection is a combination of the receiver's location on the cylinder, and a spherical transform of a ray centered at the receiver:
\begin{align}
\begin{bmatrix}
x\\y\\z
\end{bmatrix}
&= \Pi^{-1}(
\begin{bmatrix}
    u\\ v \\ r
\end{bmatrix})
=
\begin{bmatrix}
r \cos \theta(u, v) \cos \phi(v) + r_0\cos \frac{2\pi u}{W}\\
r \sin \theta(u, v) \cos \phi(v) + r_0\sin \frac{2\pi u}{W}\\
r \sin \phi(v)\\
\end{bmatrix},\label{eq:ouster-practical}
\\
\theta(u, v) &= \frac{2\pi u}{W} + \thetalut[v] \label{eq:azimuth-offset},~~~~ \phi(v) = \philut[v],
\end{align}
where $\theta(u, v)$ converts the column index $u$ to the azimuth with a linear transform by the ray's horizontal offset $\thetalut[v]$.
$\phi(v)$ directly reads the elevation from $\philut$.
As the LUTs and image size are predefined, pixel-wise LUTs can be further constructed by reorganizing Eq.~\ref{eq:ouster-practical} as a pixel-wise linear function of $r$.


\subsection{Projection $\Pi$}
While the unprojection model is straightforward, its inversion is not due to the receiver offset and the non-parametric LUTs. Assuming $r_0 \llesser r$ in Eq.~\ref{eq:ouster-practical}, we obtain an approximation:
\begin{align}
\begin{bmatrix}
    r \\
    \phi \\
    \theta
\end{bmatrix}
\approx 
\begin{bmatrix}
    \sqrt{x^2 + y^2 + z^2} \\
    \arcsin \frac{z}{r} \\
    \arctan \frac{y}{x}
\end{bmatrix},
\end{align}
then get $\hat{u} = \frac{W}{2\pi} \theta$\footnote{A warp to $[0, 2\pi]$ from $[-\pi, \pi]$ is required when using \texttt{arctan2}.} by temporarily omitting the offset $\thetalut[v]$. We then compensate $x$ and $y$ from the $r_0$ offset in Eq.~\ref{eq:ouster-practical} with approximated $\hat{u}$ and repeat the estimate until convergence. 
%
%
With the known LUT $\philut$, we then search $v$ by
\begin{align}
v &= \argmin_{t\in\{0,1,\cdots, H-1\}}~ \big\lVert \philut[t] - \phi \big\rVert.
\end{align}
To speed up the process, we construct an inverse LUT $\philut^{-1}$ with a predefined resolution, and apply $v = \philut^{-1}(\phi)$. In practice, for a $\philut$ with $H$ entries, a $\philut^{-1}$ with $2H$ entries ensures the elevation index error bounded by $\pm 1$. 
Given the estimated $v$, we finally obtain $u$ by reverting Eq.~\ref{eq:azimuth-offset}: $u = \hat{u} - \frac{W}{2\pi} \thetalut[v]$. A chain of aforementioned operations form the imperative projection function $\begin{bmatrix} u, v, r \end{bmatrix}^\top = \Pi(\begin{bmatrix}x, y, z\end{bmatrix}^\top)$.
To our best knowledge, the consideration of pixel-wise ray offset has not been presented so far in previous LiDAR unprojection and projection on range images.
Note that for the popular LiDAR dataset presented in unstructured point clouds without intrinsics, such as KITTI~\cite{Geiger2012CVPR}, our model reduces to synthetic projection~\cite{Behley2018EfficientSS} with $\thetalut(\cdot) = 0$ and $\philut(\phi) = H \cdot \frac{\phi_{\max} - \phi}{\phi_{\max} - \phi_{\min}}$, where $(\phi_{\min}, \phi_{\max})$ indicate the sensor's field of view.

\section{Registration and Surface Reconstruction}
\label{sec:method}
\begin{figure}[t]
    \centering
    \begin{tabular}{@{}c@{\hspace{1mm}}c@{\hspace{1mm}}c@{\hspace{1mm}}c}
    \includegraphics[width=.24\columnwidth,trim={15cm 0 15cm 0},clip]{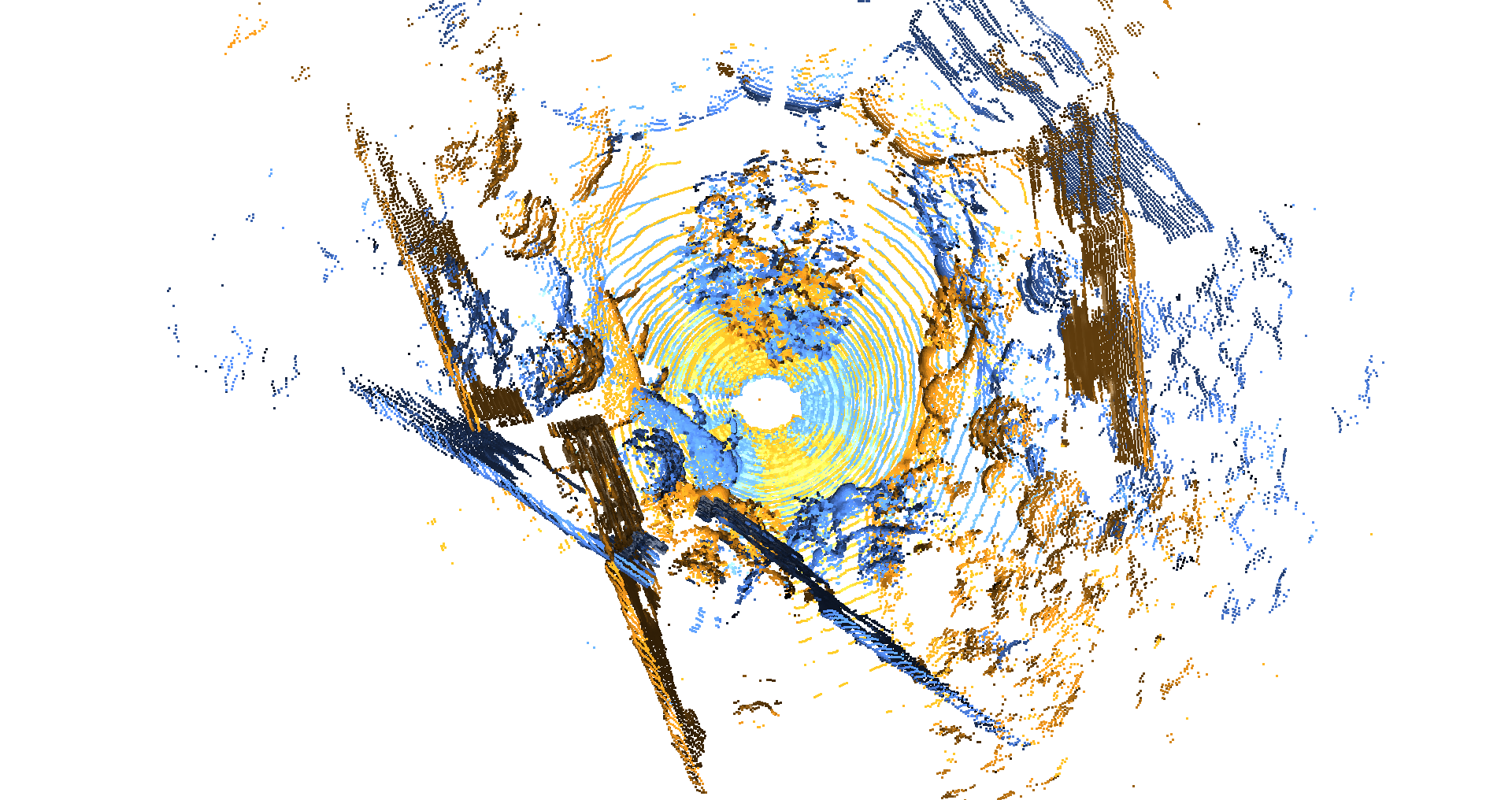} &
    \includegraphics[width=.24\columnwidth,trim={15cm 0 15cm 0},clip]{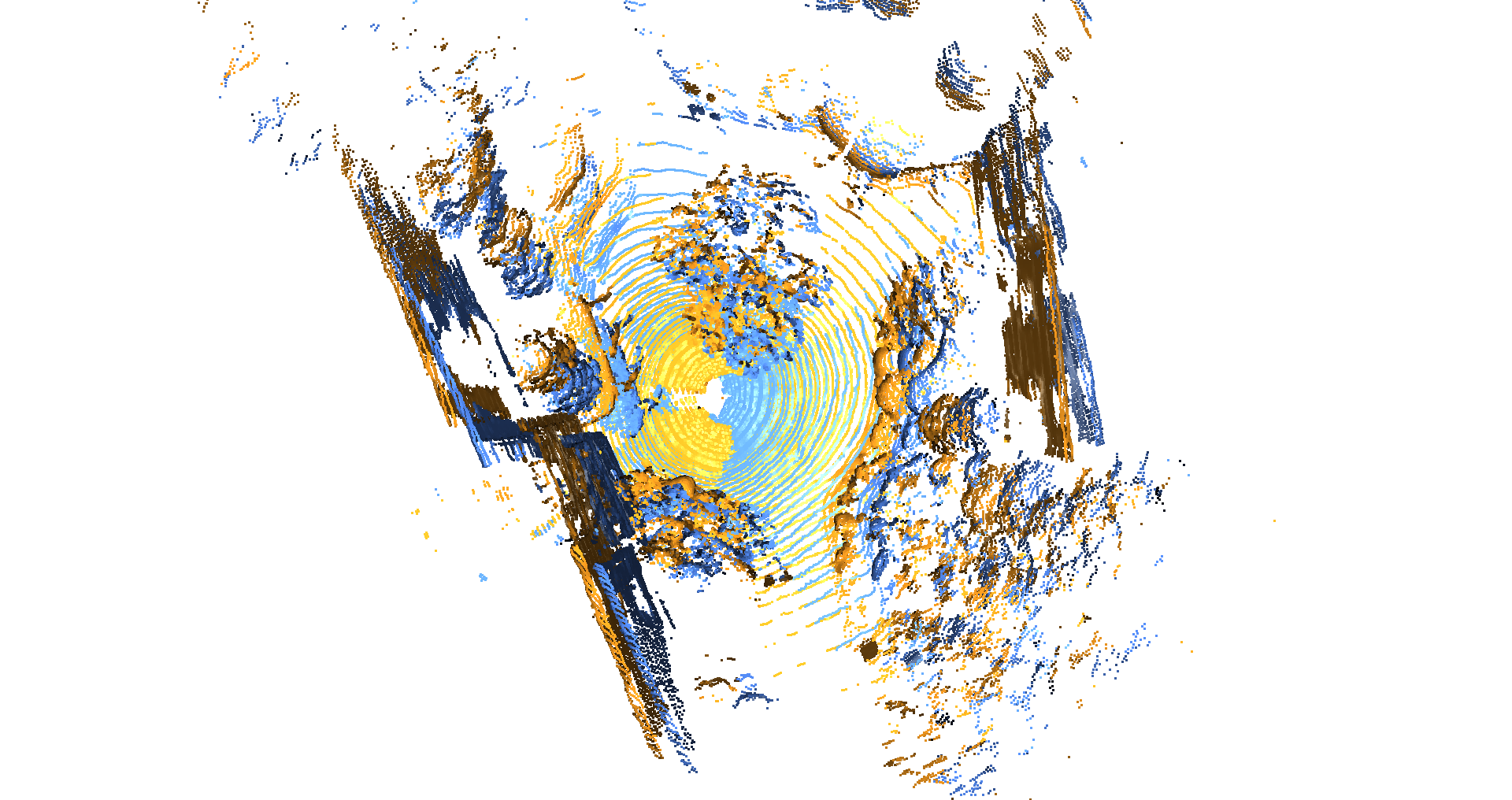} &
    \includegraphics[width=.24\columnwidth,trim={15cm 0 15cm 0},clip]{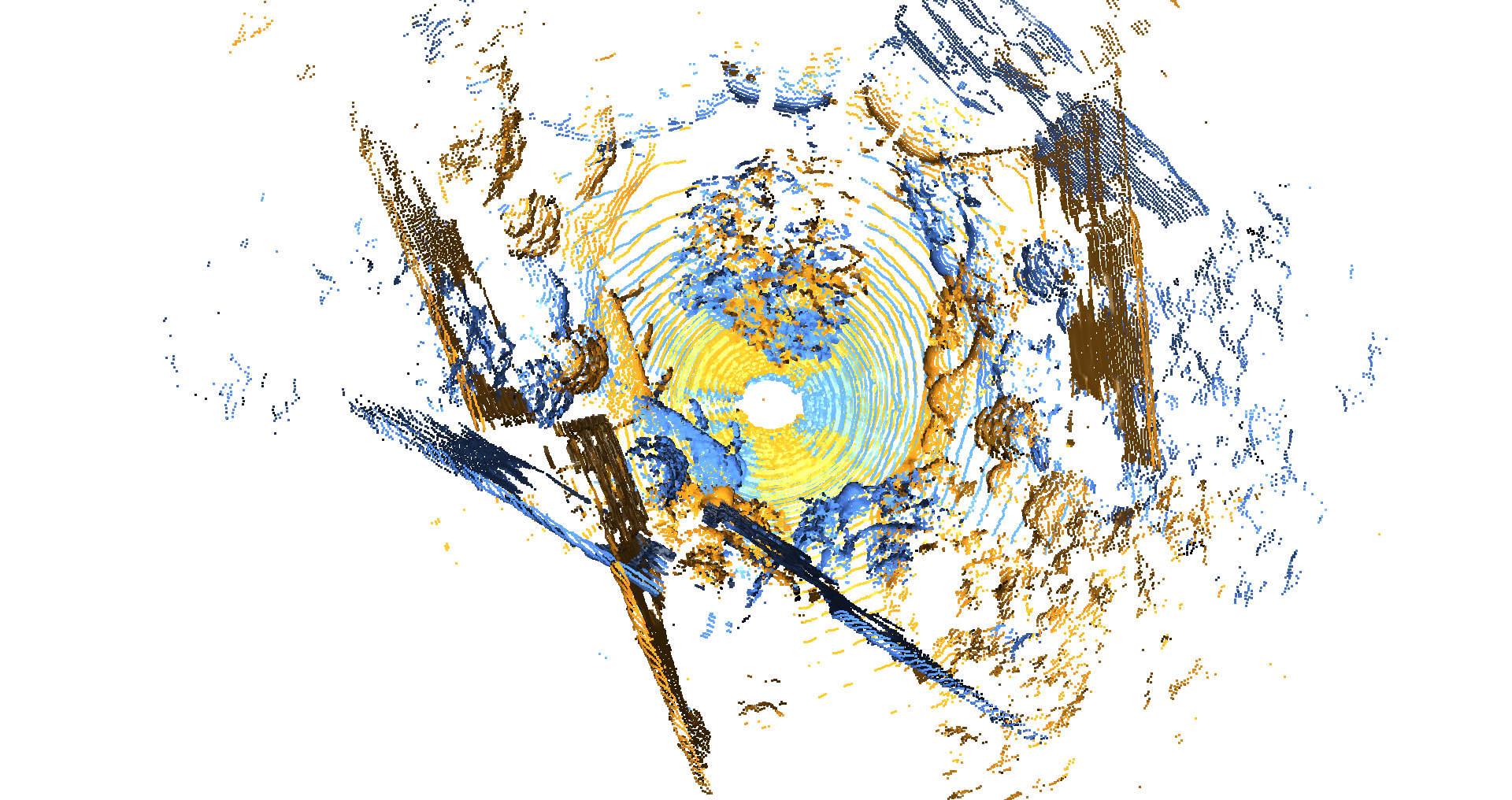} &
    \includegraphics[width=.24\columnwidth,trim={15cm 0 15cm 0},clip]{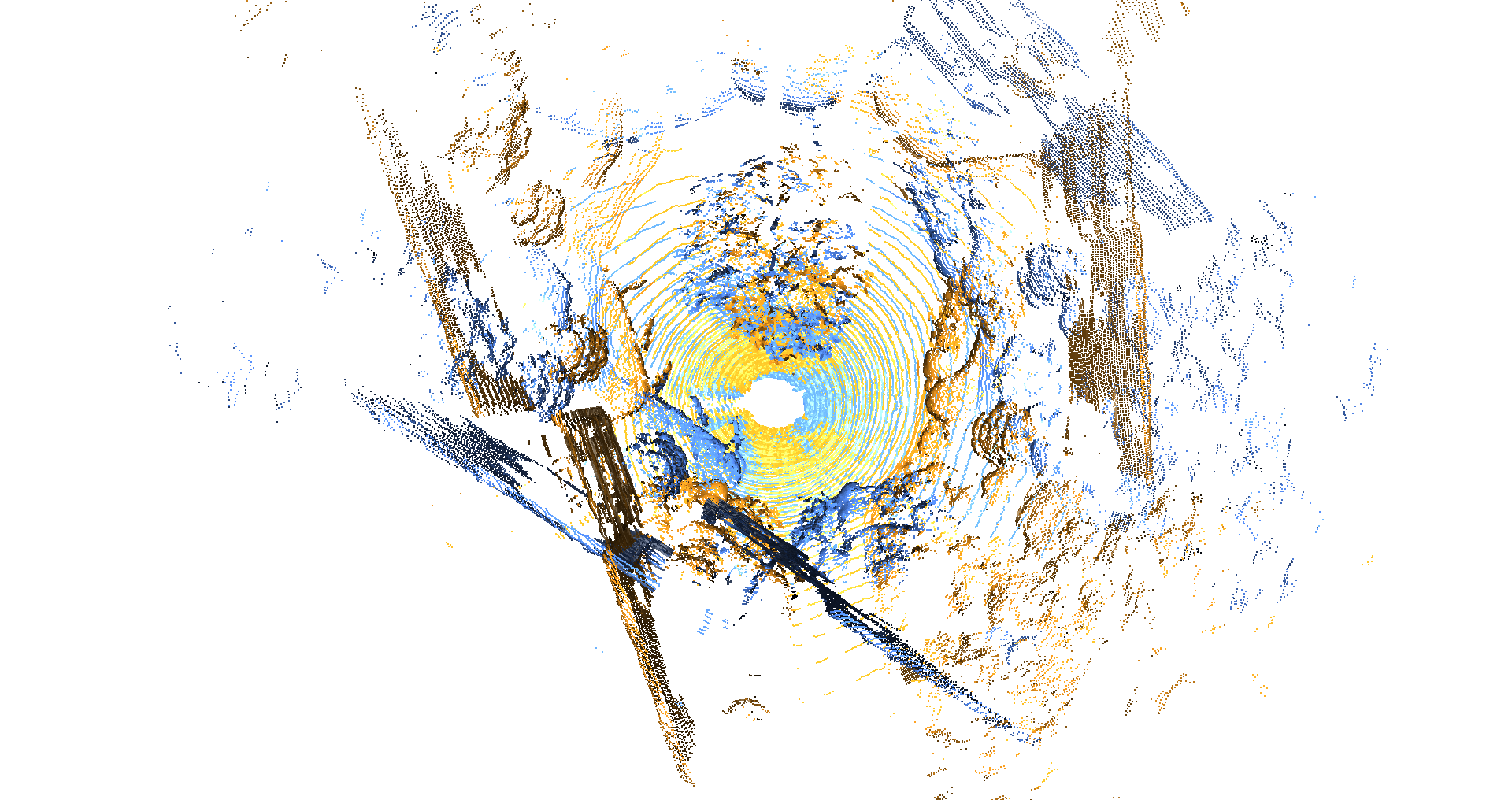} \\
    \ssf{(a) Point2Plane~\cite{Zhou2018}} & \ssf{(c) FGR~\cite{Zhou2016}} & \ssf{(e) Synthetic} & \ssf{(g) LUT} \\
    \includegraphics[width=.24\columnwidth,trim={15cm 0 15cm 0},clip]{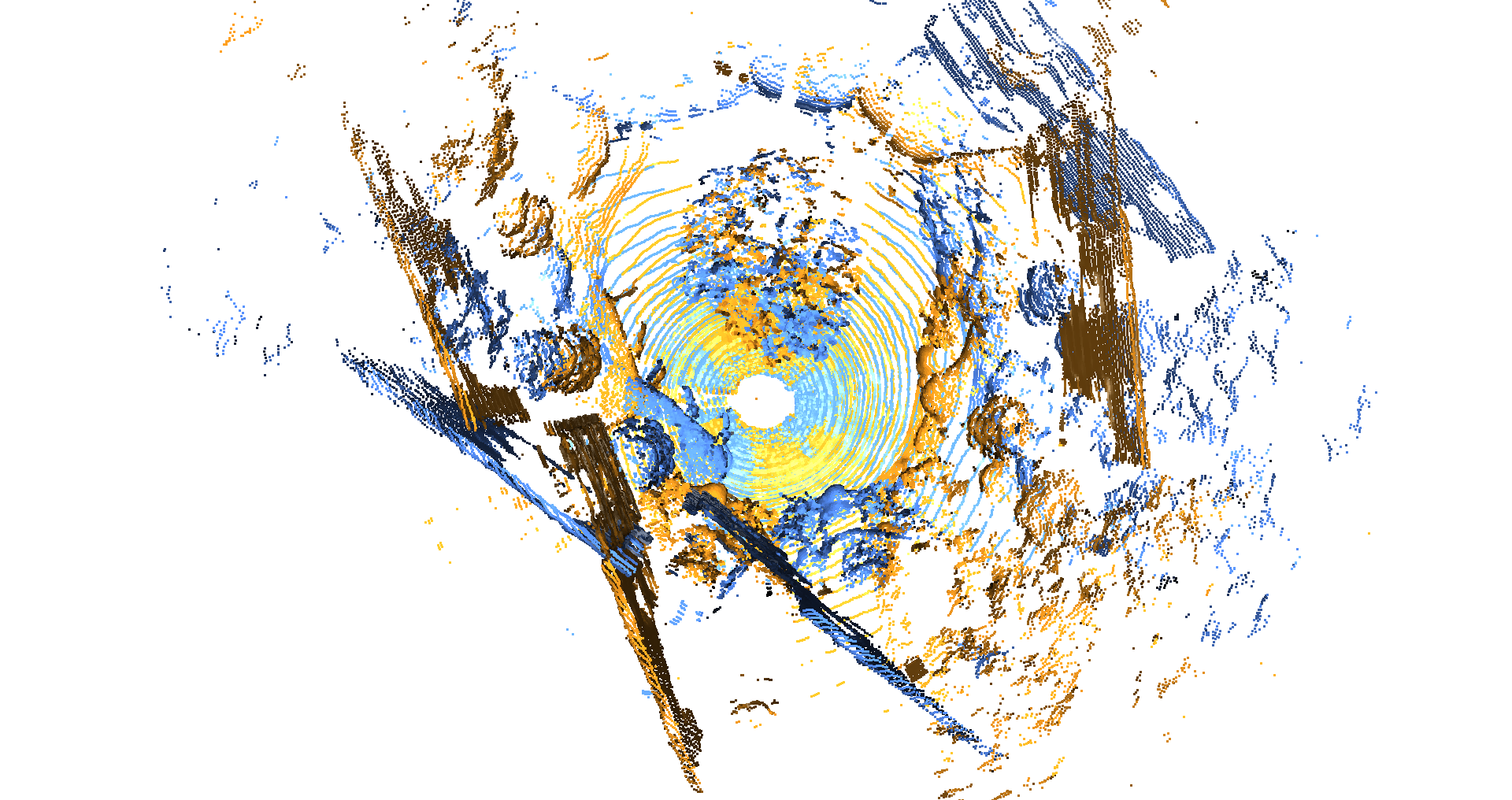} &
    \includegraphics[width=.24\columnwidth,trim={15cm 0 15cm 0},clip]{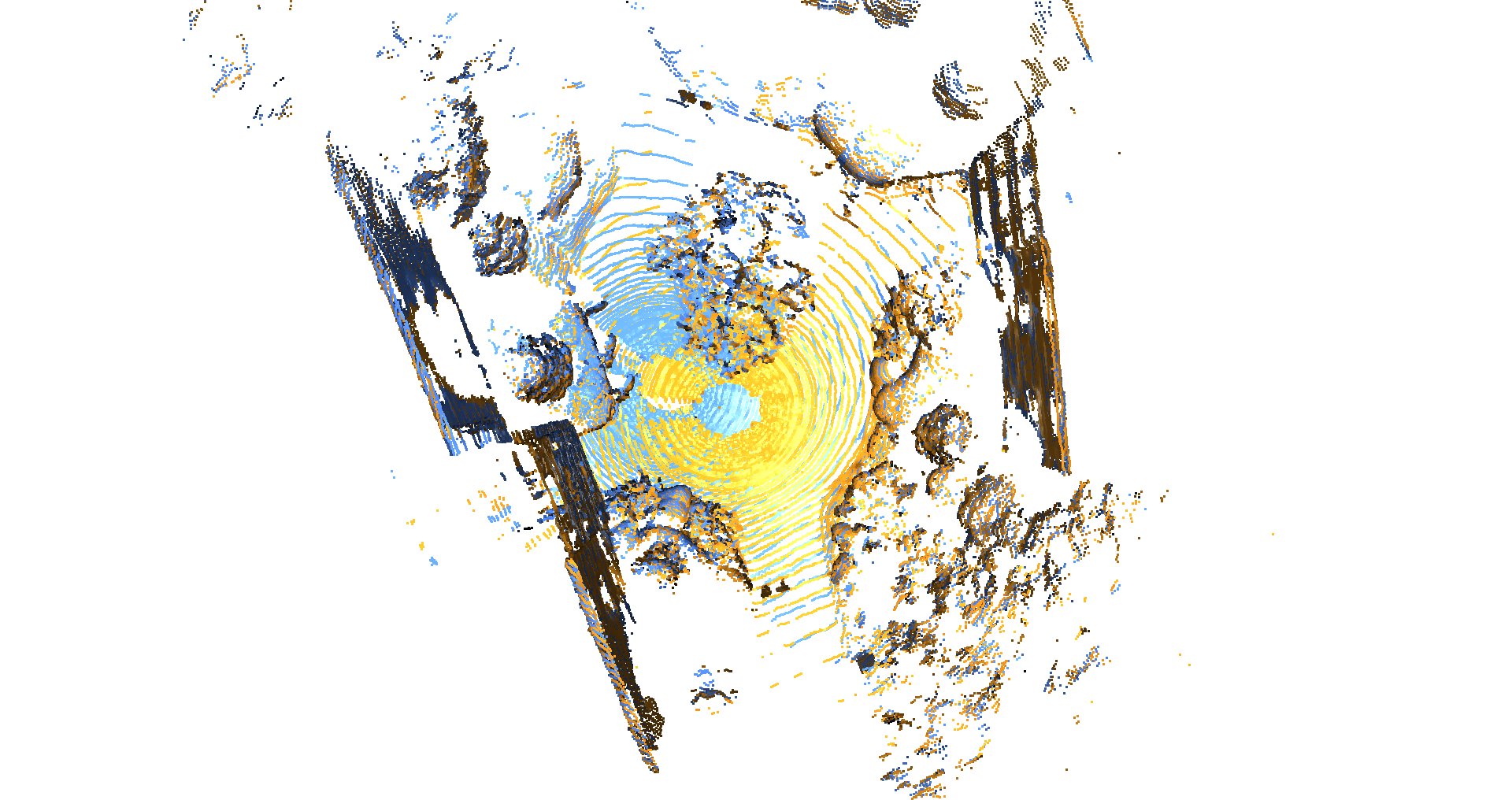} &
    \includegraphics[width=.24\columnwidth,trim={15cm 0 15cm 0},clip]{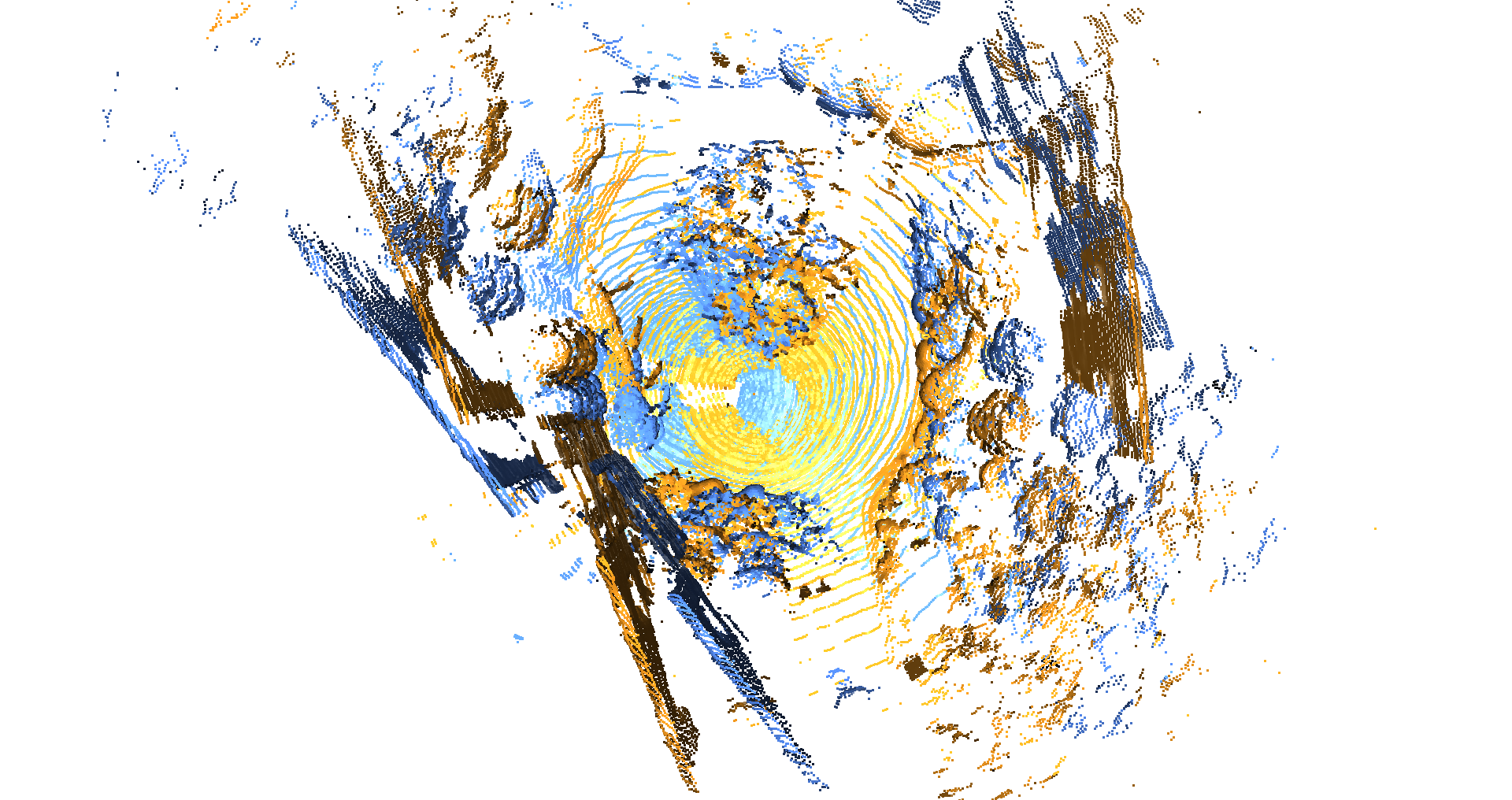} &
    \includegraphics[width=.24\columnwidth,trim={15cm 0 15cm 0},clip]{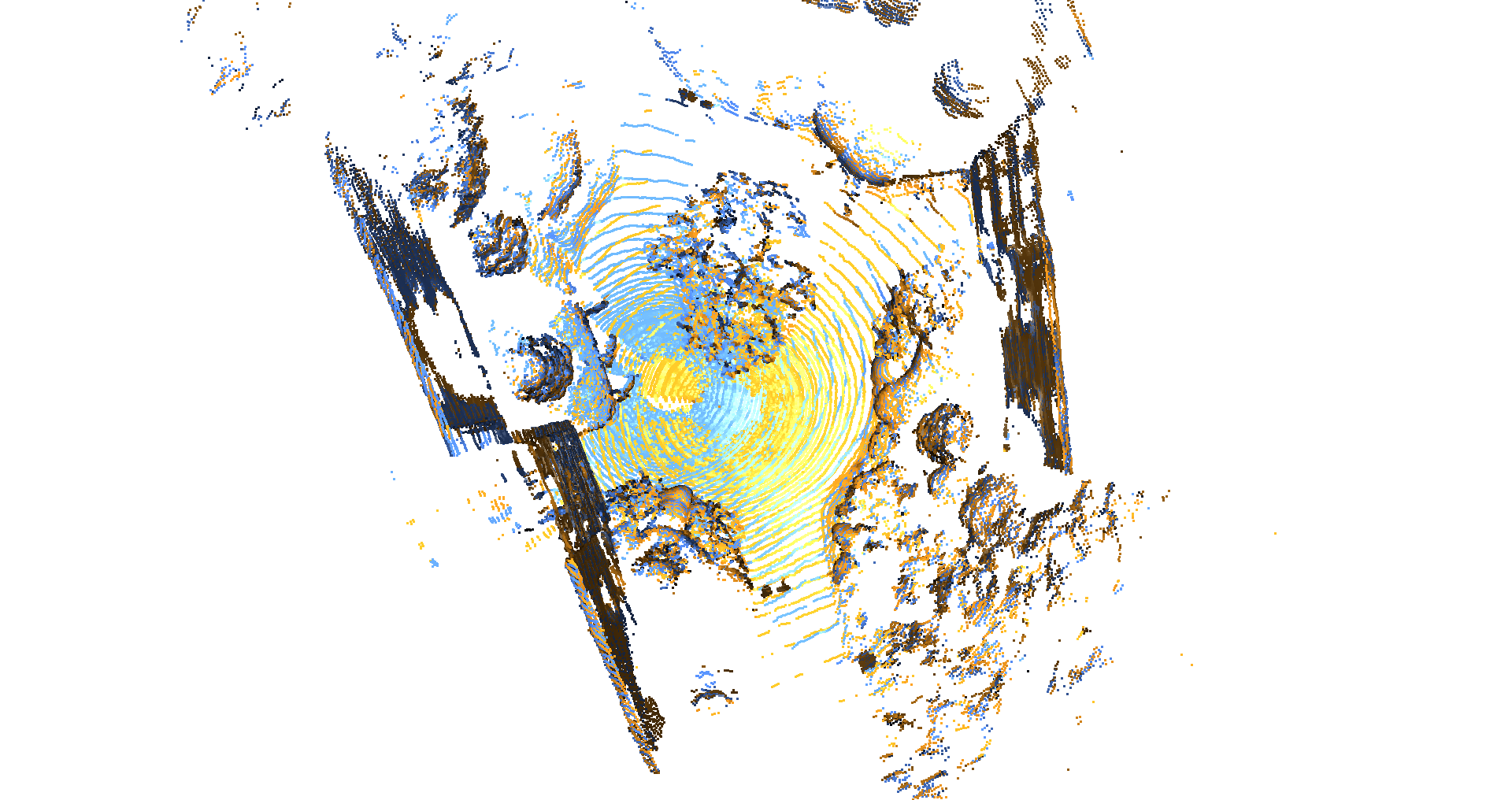} \\
    \ssf{(b) G-ICP~\cite{segal2009generalized}} & \ssf{(d) RANSAC~\cite{rusu2009fast}} & \ssf{(f) Synthetic-Multi} & \ssf{(h) LUT-Multi}
    \end{tabular}
    \caption{Illustration of registration of a challenging pair in the \emph{dormitory} sequence. While point cloud based ICP variants fail, multi-scale projective range image registration has a better convergence, and achieves comparable performance to global registration methods with known intrinsic LUTs.}\label{fig:postech-odometry}
    \vspace{-3mm}
\end{figure}

\begin{figure*}[t]
    \centering
    \begin{tabular}{c@{\hspace{1mm}}c@{\hspace{1mm}}cc}
        \includegraphics[height=2.5cm]{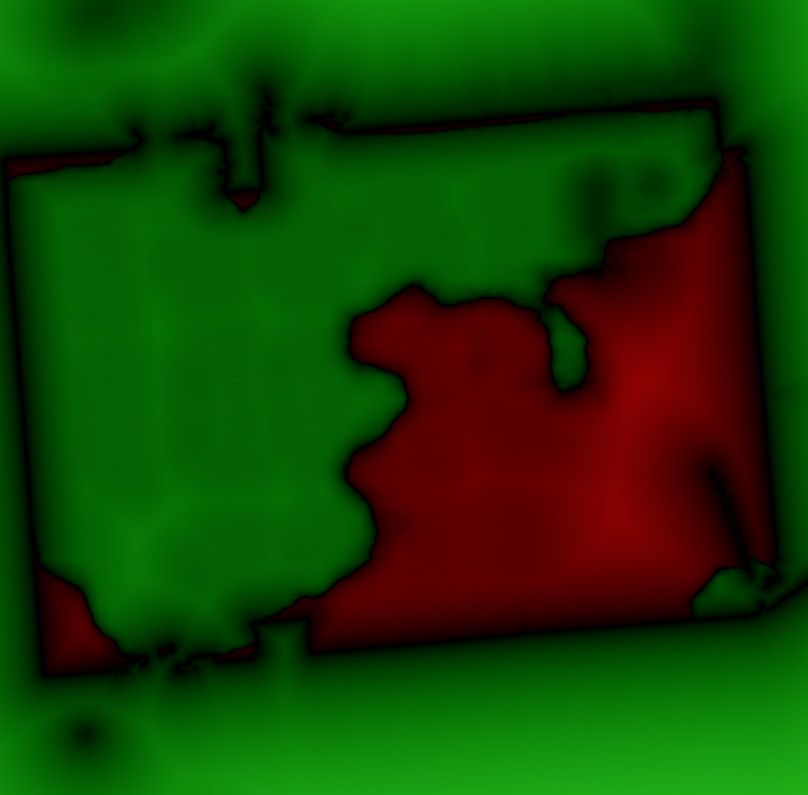} &
        \includegraphics[height=2.5cm]{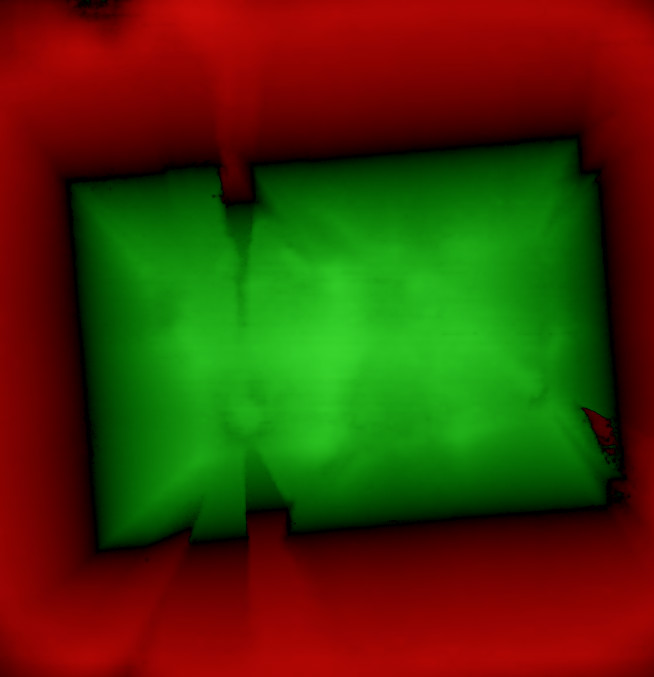} &        
        \includegraphics[height=2.5cm]{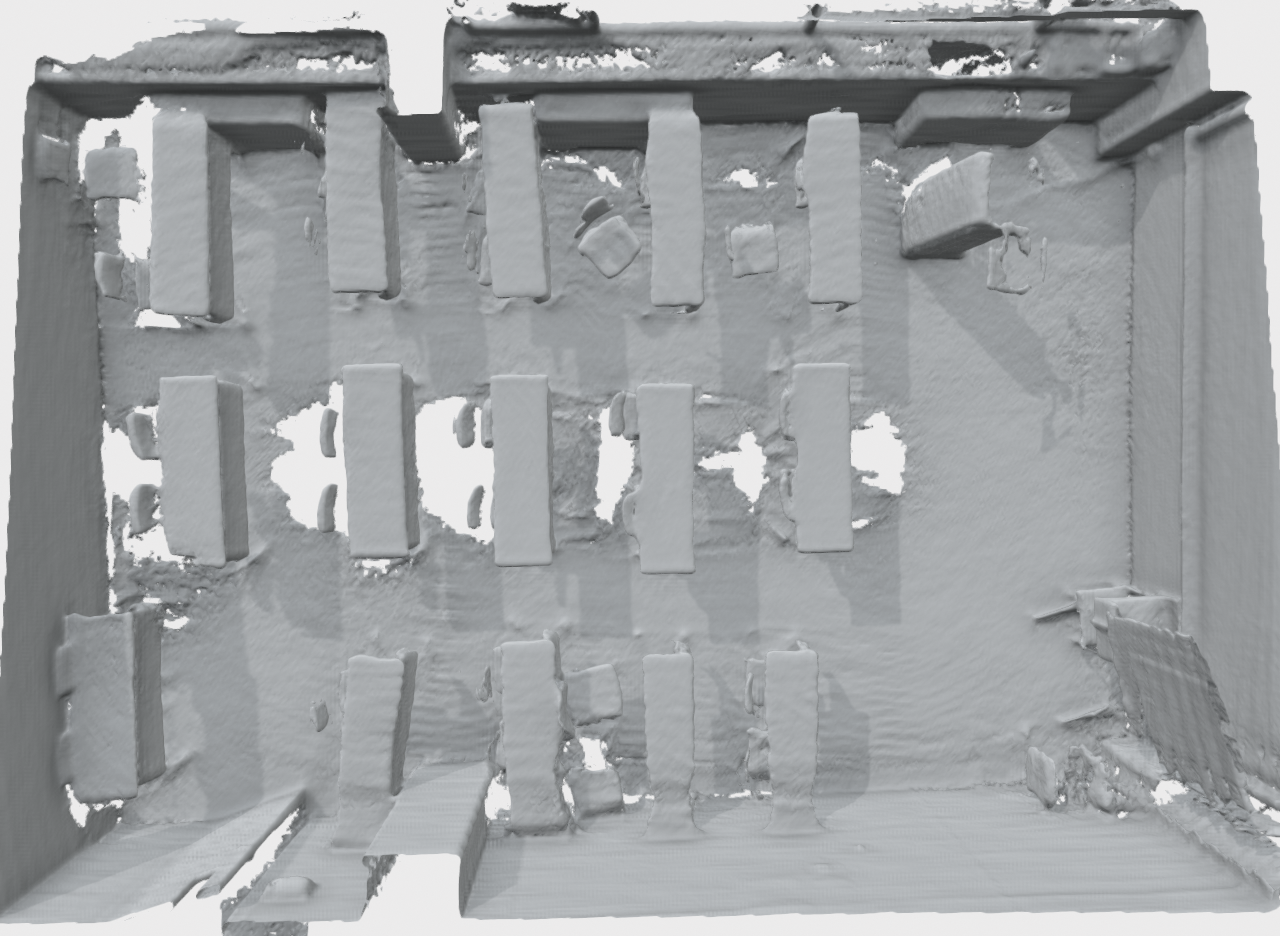} &
        \includegraphics[height=2.5cm]{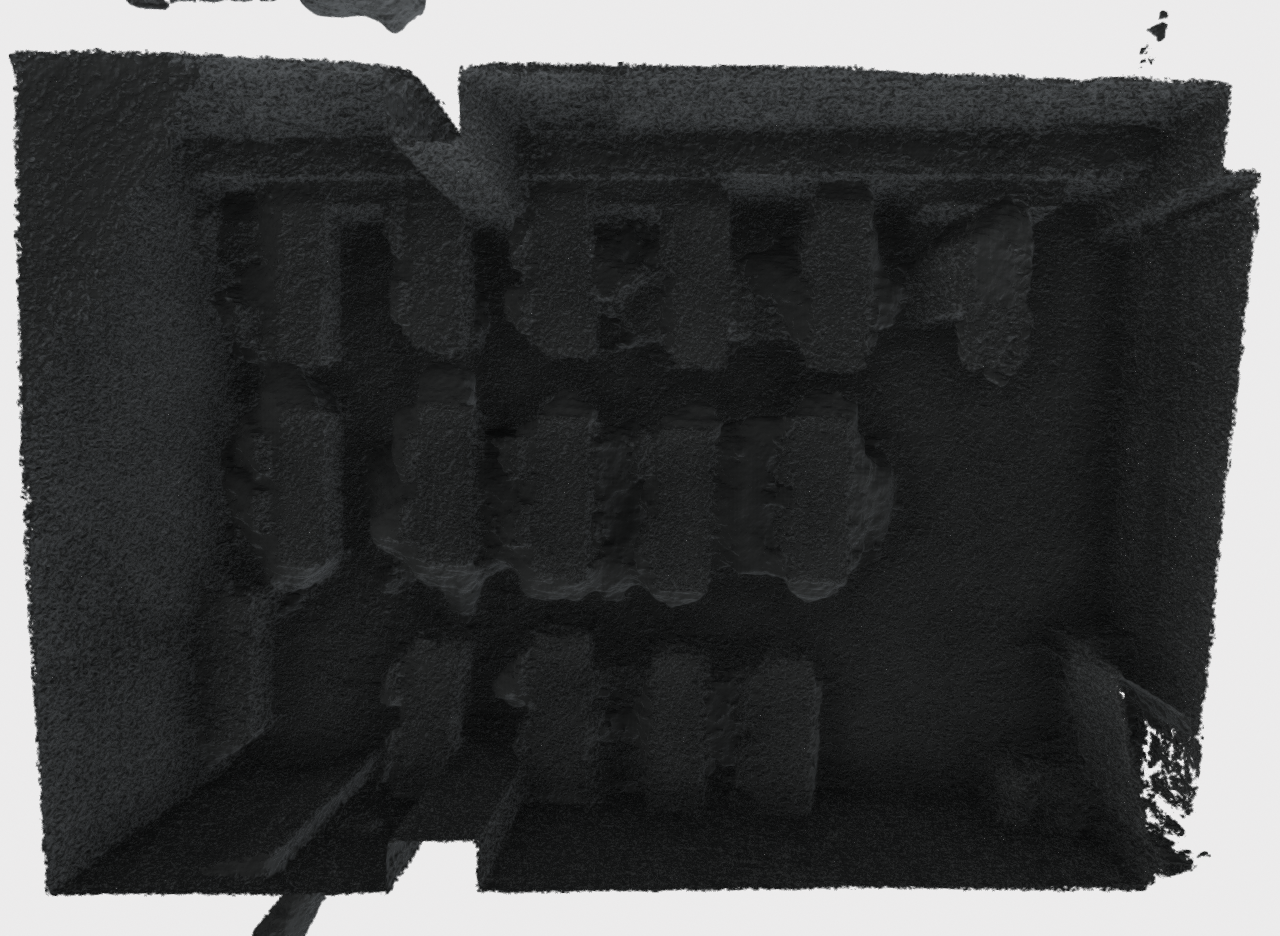} \\
        (a) & (b) & (c) & (d)
    \end{tabular}
    \caption{(a) and (b): neural SDF level sets from (a) mesh and (b) projective SDF from range images on scene \emph{lecture room}, rendered with {\sf instant-sgp}~\cite{mueller2022instant}. Green and red indicate positive and negative predictions respectively. Neural SDF designed for mesh prefers positive SDF samples, while projective SDF has distinguishing +/- half planes. (c) and (d): Reconstructed mesh via Marching Cubes from (c) discrete projective SDF volumes and (d) neural SDF trained with projective SDF. While being able to fill holes, (d) looks darker due to irregular normals from neural SDF's reduced compatibility to projective SDF.}
    \label{fig:neural-sdf}
    \vspace*{-3mm}
\end{figure*}
\vspace{-3mm}

\subsection{Multi-scale Cylindrical Range Image Registration}
We now register two LiDAR scans through index-based projective data association. 
We take the source scan in the point cloud form $\pp \in \PP_\mathrm{src}$ (by applying unprojection), and the target scan in the range image form $\Omega_{\mathrm{dst}}$.

With an initial transformation $\RR_k \in SO(3), \tt_k \in \RR^3$ (typically $\RR_k$ initialized to identity and $\tt_k$ estimated by aligning two point set centers), we get the associated point cloud $\qq \in \QQ_{\mathrm{dst}}$ by\footnote{\label{note-clarity} Image boundary check is ignored for clarity. Same for SDF reconstruction.}
\begin{align}
    [u, v, r]^\top &= \Pi(\RR_k \pp + \tt_k), \label{eq:assoc_proj}\\
    \qq &= \Pi^{-1}\bigg(u, v, \Omega_{\mathrm{dst}}(u, v)\bigg) \label{eq:assoc_unproj},
\end{align}
where $u, v, r$ are pixel coordinates and range, and $\Omega_{\mathrm{dst}}$ reads the range measurements at $(u, v)$. These operations can be easily vectorized and run in parallel.
Denote the correspondence set with $\cC = \{(\pp_i, \qq_j) \mid \pp_i \in \PP_{\mathrm{src}},~ \qq_j \in \QQ_{\mathrm{dst}}\}$, we have the nonlinear least squares estimate using Gauss-Newton from
\begin{align}
    \RR_{k+1}, \tt_{k+1} = \argmin_{\RR, \tt} \sum_{{\pp_i, \qq_j} \in \cC} \rho \bigg(\lL (\RR \pp_i + \tt, \qq_j)\bigg), \label{eq:assoc_solve}
\end{align}
where $\lL$ is the point-to-plane loss $\lL(\xx, \yy) = \nn_\yy^\top (\xx - \yy)$ given the normal $\nn_\yy$, attached with a robust kernel $\rho$~\cite{barron2019general}. The normal image can be efficiently constructed by eigenvalue decomposition of nearest neighbors in a searching window, or simply a cross product of two neighbor pixels~\cite{Behley2018EfficientSS}. Iterating Eqs.~\ref{eq:assoc_proj}-\ref{eq:assoc_solve} constructs the range image based registration algorithm. Implementation-wise, the projective data assciation discards the use of a k-d tree that requires $O(N \log N)$ construction and query time in two passes, therefore the $O(N)$ correspondence search and linear system construction can be finished in one pass in parallel.

While the cylindrical range image has a wide receptive field in the horizontal direction, putative correspondences $\cC$ are still limited. In view of this, we propose \emph{multi-scale registration} for the task. A range image pyramid is constructed by accessing strided range and normal images, retaining the original LiDAR intrinsics. Projective transforms are performed at the finest level, but down-sampled on coordinates at the given stride. Compared to point clouds, image-based downsampling takes no time by only changing the strides in the projection model, and does not need voxelization of point clouds that requires the $O(N)$ construction of a spatial hash map. 

As a result, multi-scale registration for cylindrical images significantly boosts fidelity of registration, and it lifts the local registration algorithm in the ICP-fashion to be comparable to global registration approaches such as RANSAC. Fig.~\ref{fig:postech-odometry} and Fig.~\ref{fig:kitti-odometry} show registration examples.

\begin{figure*}[t]
\centering
\begin{tabular}{@{}c@{\hspace{1mm}}c@{\hspace{1mm}}c@{}}
    \includegraphics[width=0.32\textwidth]{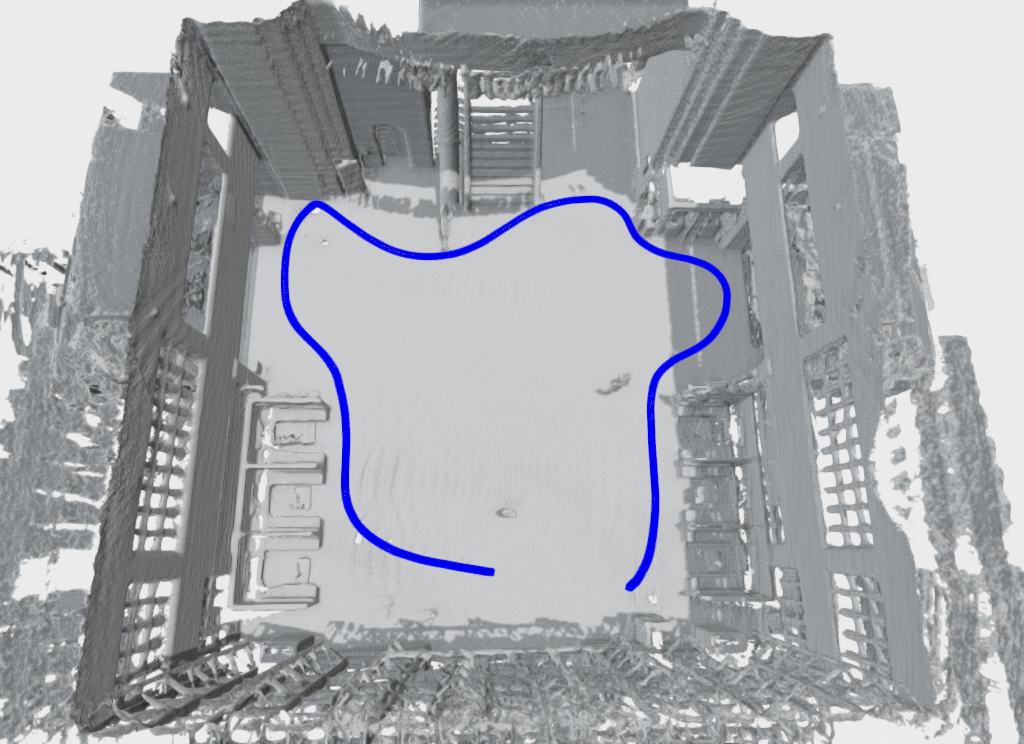} & 
    \includegraphics[width=0.32\textwidth]{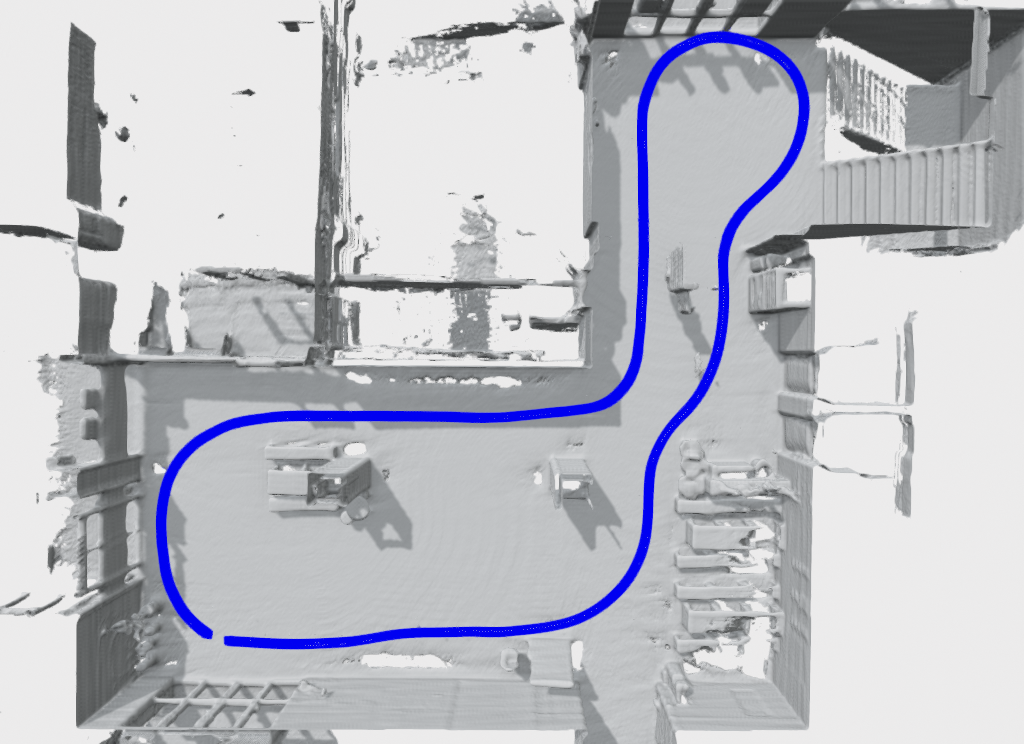} &     
    \includegraphics[width=0.32\textwidth]{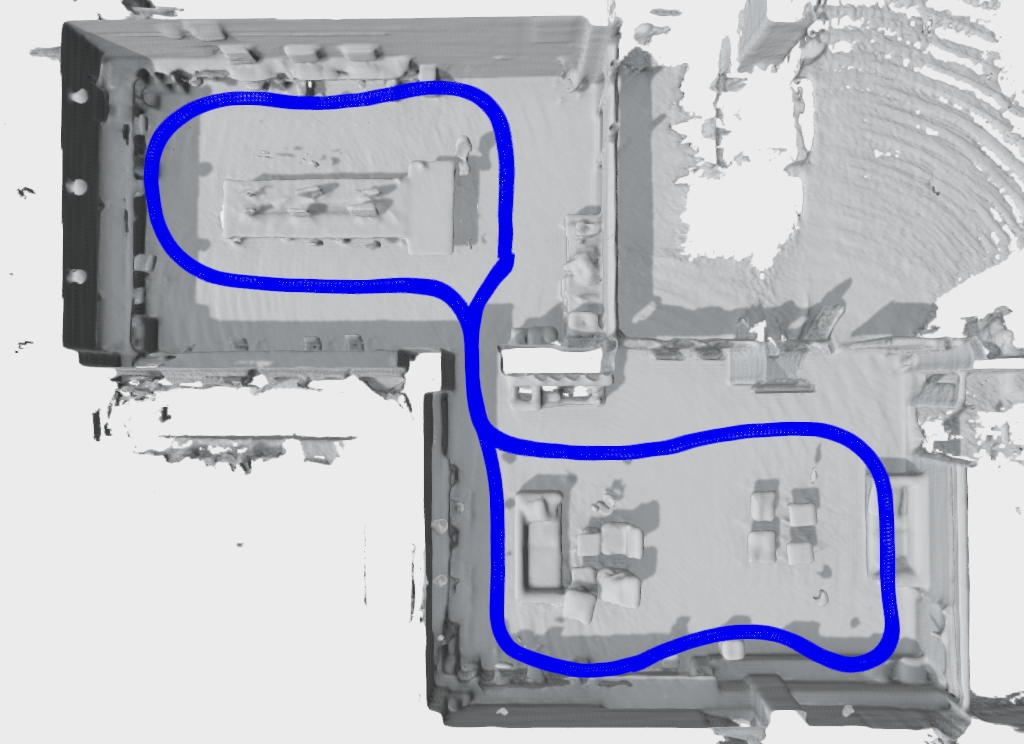} \\        
    {\small(a) Students room} & {\small (b) Lecture building} & {\small (c) Lounge}\\
    \includegraphics[width=0.32\textwidth]{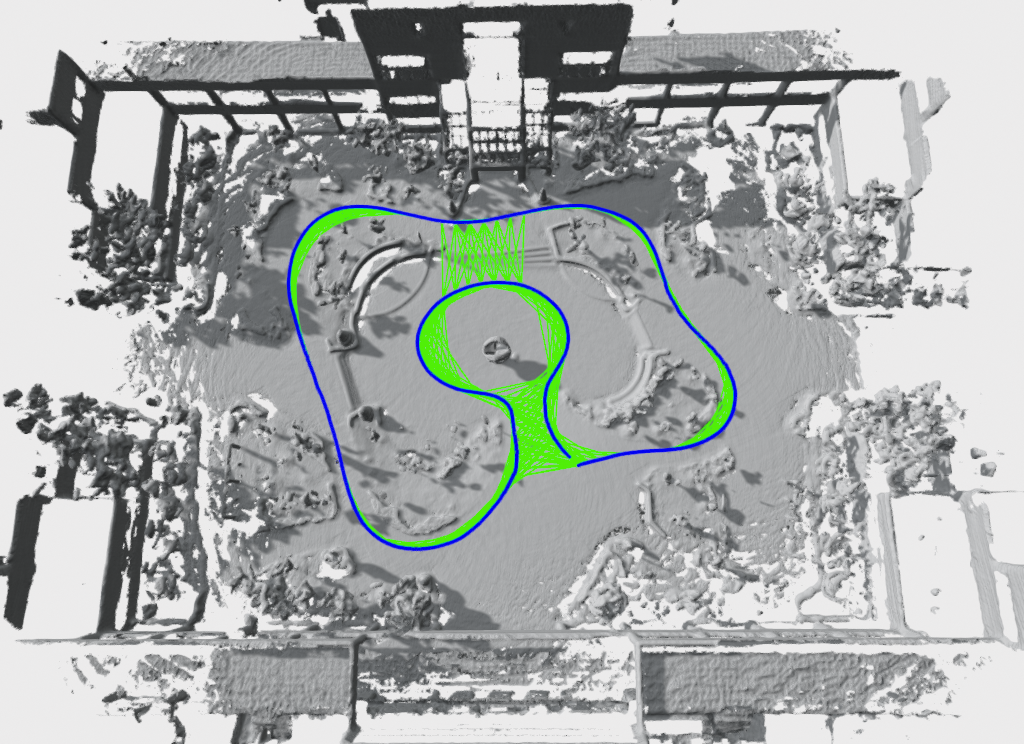} & 
    \includegraphics[width=0.32\textwidth]{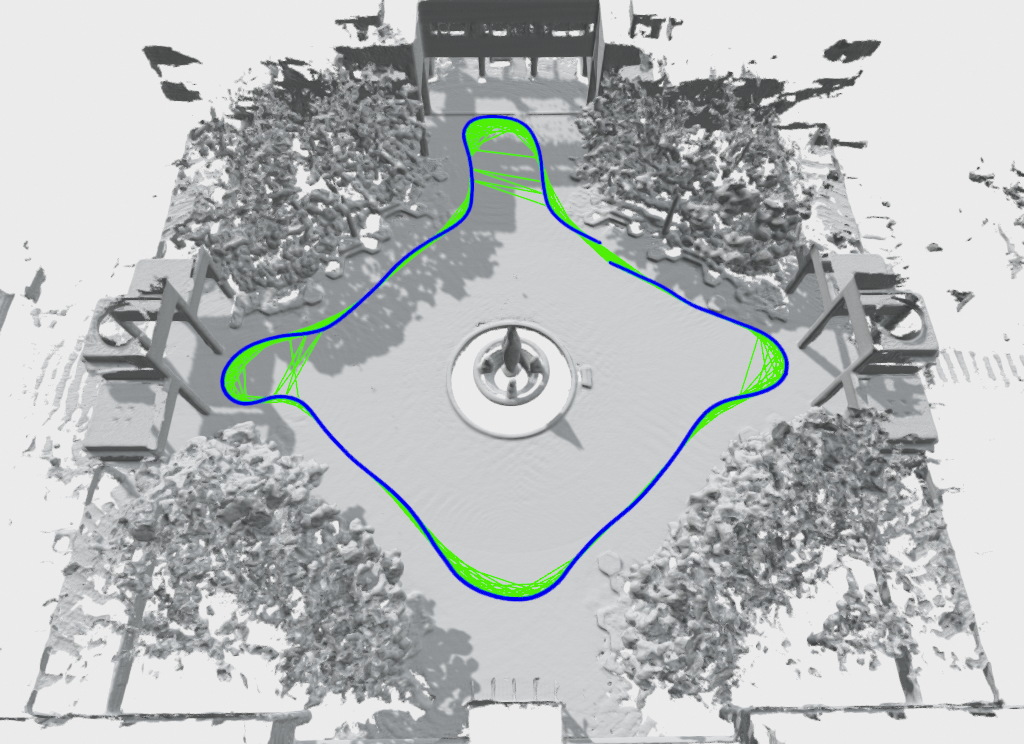} &     
    \includegraphics[width=0.32\textwidth]{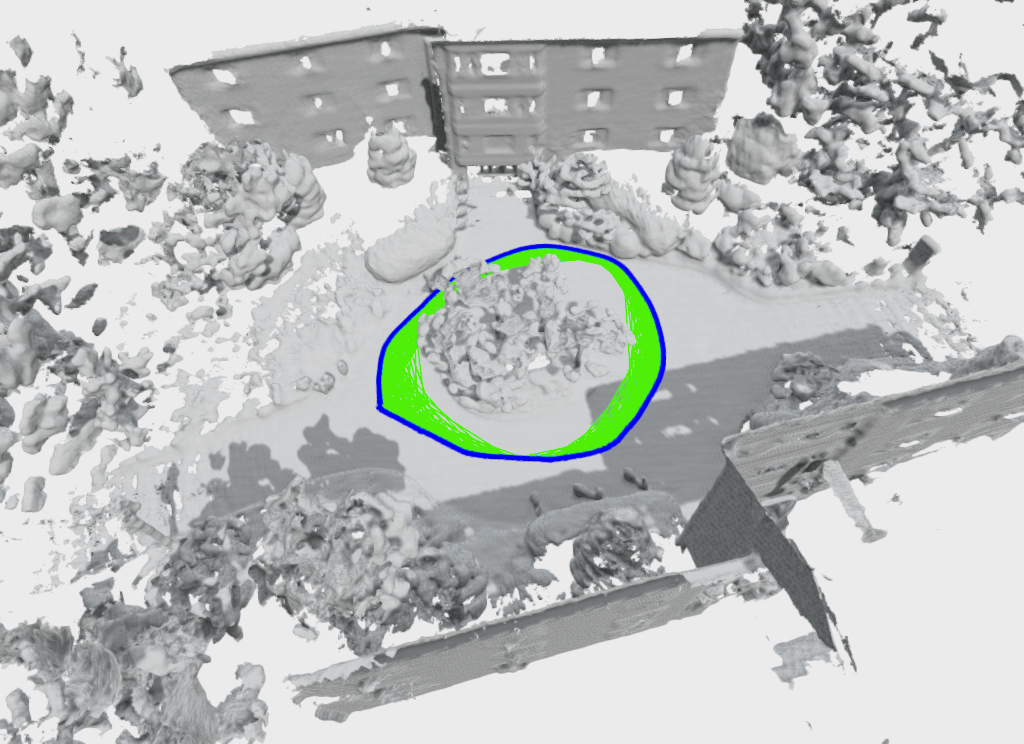} \\    
    {\small(d) Square} & {\small(e) Fountain} & {\small(f) Dormitory} \\
\end{tabular}
\caption{Surface reconstruction via SDF integration of selected sequences from our dataset, overlaid with sensor trajectory (poses in blue, loop closures in green). Top: indoor scenes. Bottom: outdoor scenes. Loop closures for indoor scenes are omitted to avoid occlusion of geometry details. }
\label{fig:dataset}
\vspace*{-3mm}
\end{figure*}
\subsection{Signed Distance Function from LiDAR Range Images}
One of the key benefits of using range images for LiDARs is that we can naturally apply parallel SDF estimation for dense surface reconstruction.
SDF measures the distance from an arbitrary query point to its nearest surface. With a perfect watertight mesh model, signed distance per point can be computed via ray casting~\cite{takikawa2021nglod}. In real world with accumulating data, such computation is intractable especially for online usages.
For LiDAR data, a common practice is to cast rays from sensor origin to scan points and update samples along the ray. This formation, however, limits the sampling distribution, and is not friendly to parallel computation due to race conditions at ray intersections.
Classical volumetric reconstruction~\cite{curless1996volumetric, newcombe2011kinectfusion, niessner2013real} projects arbitrary 3D points to depth images and computes weight average of truncated projective SDF to approximate the real SDF. It requires a range image and a projection model where our representation fits.

To estimate the projective SDF from a query point $\xx \in \mathbb{R}^{3}$, we find its projective association in a range image~$\Omega_j$ with pose $\RR_j \in SO(3), \tt_j \in \mathbb{R}^3$, and estimate the signed distance $d_j(\xx)$ along the projection ray:
\begin{align}
    \begin{bmatrix}u & v & r\end{bmatrix}^\top &= \Pi(\RR_j \xx + \tt_j),\\
    d_j(\xx) &= \Omega_j(u, v) - r.
\end{align}
With a sequence of LiDAR range measurements $\{\Omega_{j=1}^N\}$ and their associated poses, we can get a least squares estimate at query points, typically at discretized voxel grid points:
\begin{align}
    d(\xx) &= \argmin_t \lVert t - d_j(\xx) \rVert^2 = \frac{\sum{d_j}}{N},
\end{align}
which can be updated incrementally~\cite{newcombe2011kinectfusion}. While the formulation still holds when using LiDAR projective model, LiDAR has wide range, therefore a dense grid does not scale to LiDAR range images. In this regard, we use ASH~\cite{dong2021ash} to generate a globally sparse locally dense hash grid for unbounded scene reconstruction. For each point unprojected from a range image, we activate dense voxel blocks in the shape of $16^3$ within a certain radius; only the SDF value of activated voxel blocks in the cylindrical viewing volume will be updated. Accelerated Marching Cubes~\cite{lorensen1987marching, dong2018efficient} is applied to extract a triangle mesh at zero-crossing isosurfaces.


This approximate SDF computation at arbitrary $\xx \in \mathbb{R}^3$ also opens the door to the online training of the neural SDF~\cite{takikawa2021nglod, mueller2022instant} with incremental LiDAR inputs, where a multi-layer perceptron (MLP) is trained to predict SDF value at continuous sampled positions with SDF readings. In experiments, however, we observe distinguished characteristics between SDF from mesh and range images, leading to reduced accuracy of surface prediction in the current neural rendering systems~\cite{mueller2022instant}, see Fig.~\ref{fig:neural-sdf}. We leave a full adaptation of neural SDF and surface reconstruction to range images as future work.

%

\section{A LiDAR Range Image Dataset}
\label{sec:dataset}
There has been a plethora of LiDAR datasets~\cite{Geiger2012CVPR, tan2020toronto3d, hackel2017semantic3d}, but most of them, if not all, are presented in point clouds. We therefore construct a new dataset in the range image format to fix this absence.

\subsubsection{Data Collection.}
We collect various indoor and outdoor sequences with an Ouster OS0 128 LiDAR. The selection of Ouster is the result of its user-friendly access to raw scans; an adaptation to Velodyne is also possible with low-level driver modifications.
The LiDAR is placed on a portable cart, see supplementary for details. 
A cart is a good trade-off between flexibility and stability.
It provides a stable platform that reduces vibration comparing to a hand-held setup, and is akin to the most prevalent vehicle-top setup but more flexible and works indoor. The easy-to-control motion pattern enriches registration patterns and improves scene coverage for surface reconstruction, in comparison to vehicle-top setups.
%
The outdoor sequences are collected on campus, varying from squares to dormitories. The indoor sequences are collected in buildings, ranging from halls to lecture rooms. 
All the sequences are captured in the $128 \times 1024$ resolution at 10 Hz. The sequence names are listed in Table.~\ref{tab:reconstruction} and their detailed statistics are in supplementary.

\subsubsection{Pseudo Groundtruth Pose Generation.}
To acquire poses of the range images without an available large scale motion capture system, we utilize a modified multiway registration system~\cite{Choi2015} based on Generalized ICP (G-ICP)~\cite{segal2009generalized}, 3-pt FPFH-RANSAC~\cite{fischler1981random}, and robust pose graph optimization. This setup of pseudo ground truth pose generation is common in the RGB-D datasets~\cite{zeng20163dmatch, dai2017bundlefusion, yang2021self} and has been widely used in the vision community.

For each sequence, we first apply G-ICP between adjacent frames to obtain odometry measurements and build an initial pose graph. We then select key frames every $K=10$ frames, and exhaustively apply RANSAC (max 1M iterations with confidence $0.999$) between key frames. Valid global registration results are refined with G-ICP, and inserted into the pose graph as loop closure edges. Finally, the pose graph is optimized with a robust line process~\cite{Choi2015} to filter inconsistent edges and output poses. The LiDAR scans per sequence are accumulated as a pseudo-ground truth 3D point cloud for reconstruction evaluation.
%

\section{Experiments}
\label{sec:experiment}
\begin{figure}[t]
    \centering
    \begin{tabular}{@{}c@{\hspace{1mm}}c@{\hspace{1mm}}c}
    \includegraphics[width=.32\columnwidth,trim={0 5cm 5cm 0},clip]{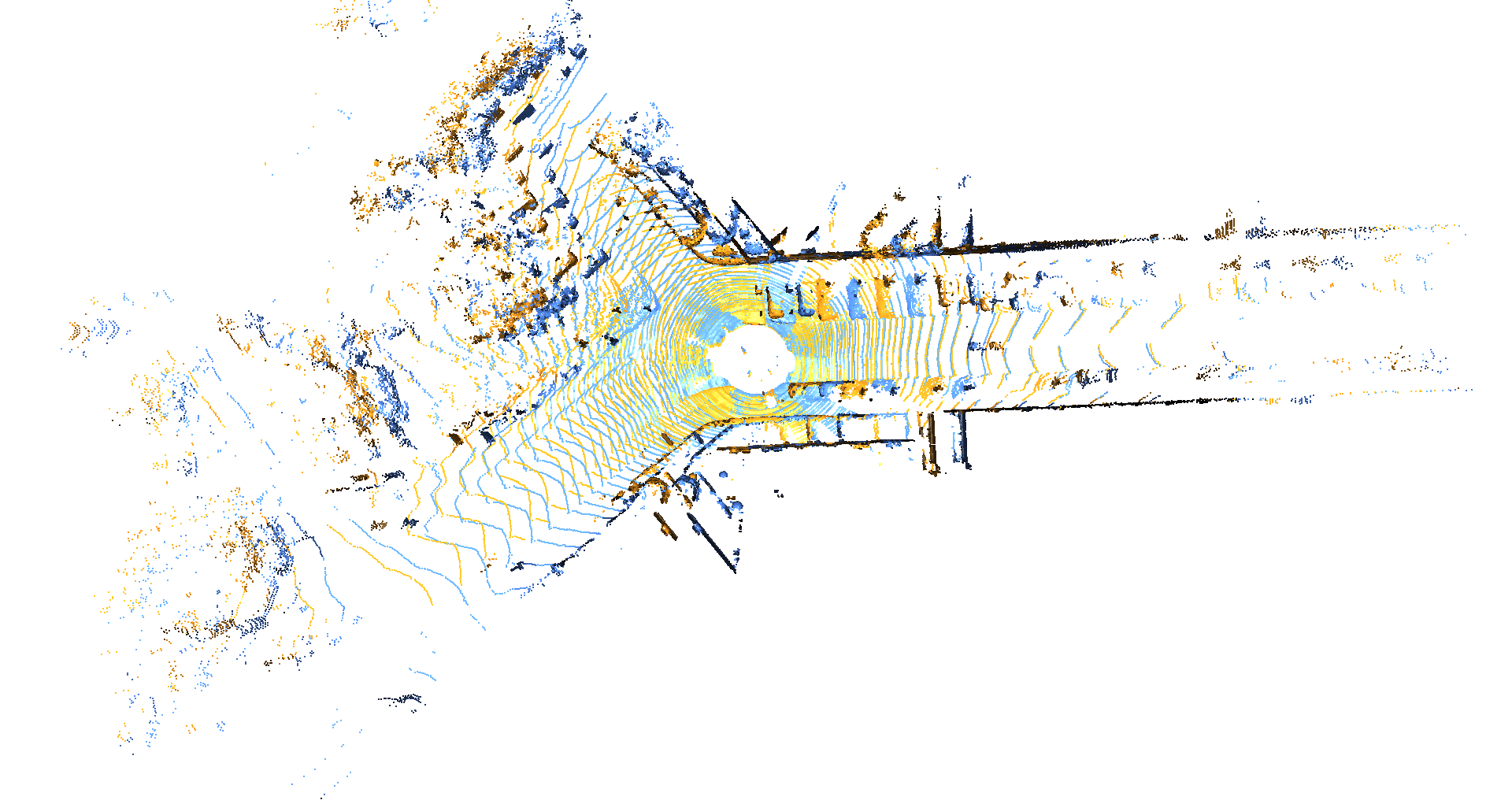} &
    \includegraphics[width=.32\columnwidth,trim={0 5cm 5cm 0},clip]{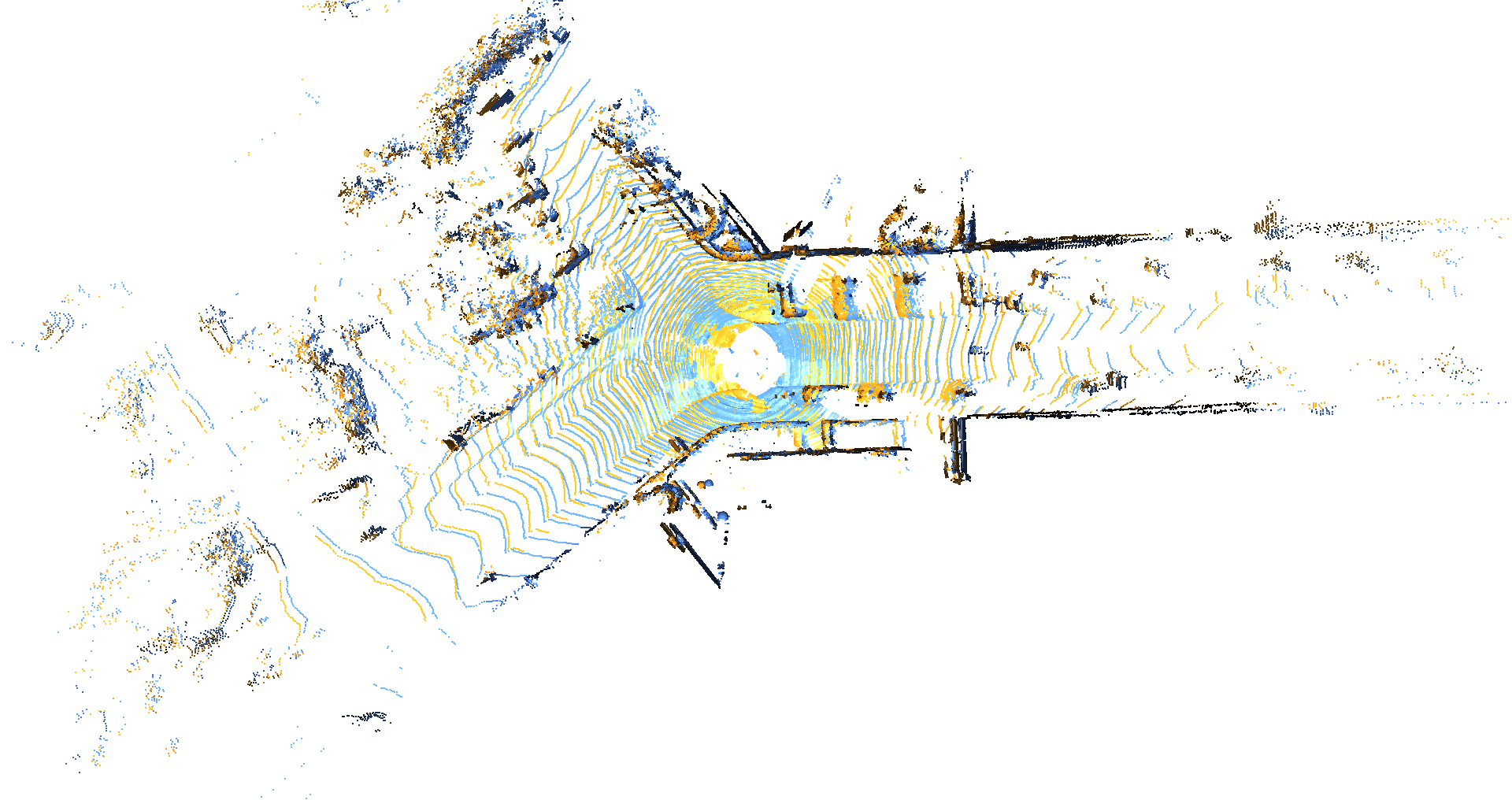} &
    \includegraphics[width=.32\columnwidth,trim={0 5cm 5cm 0},clip]{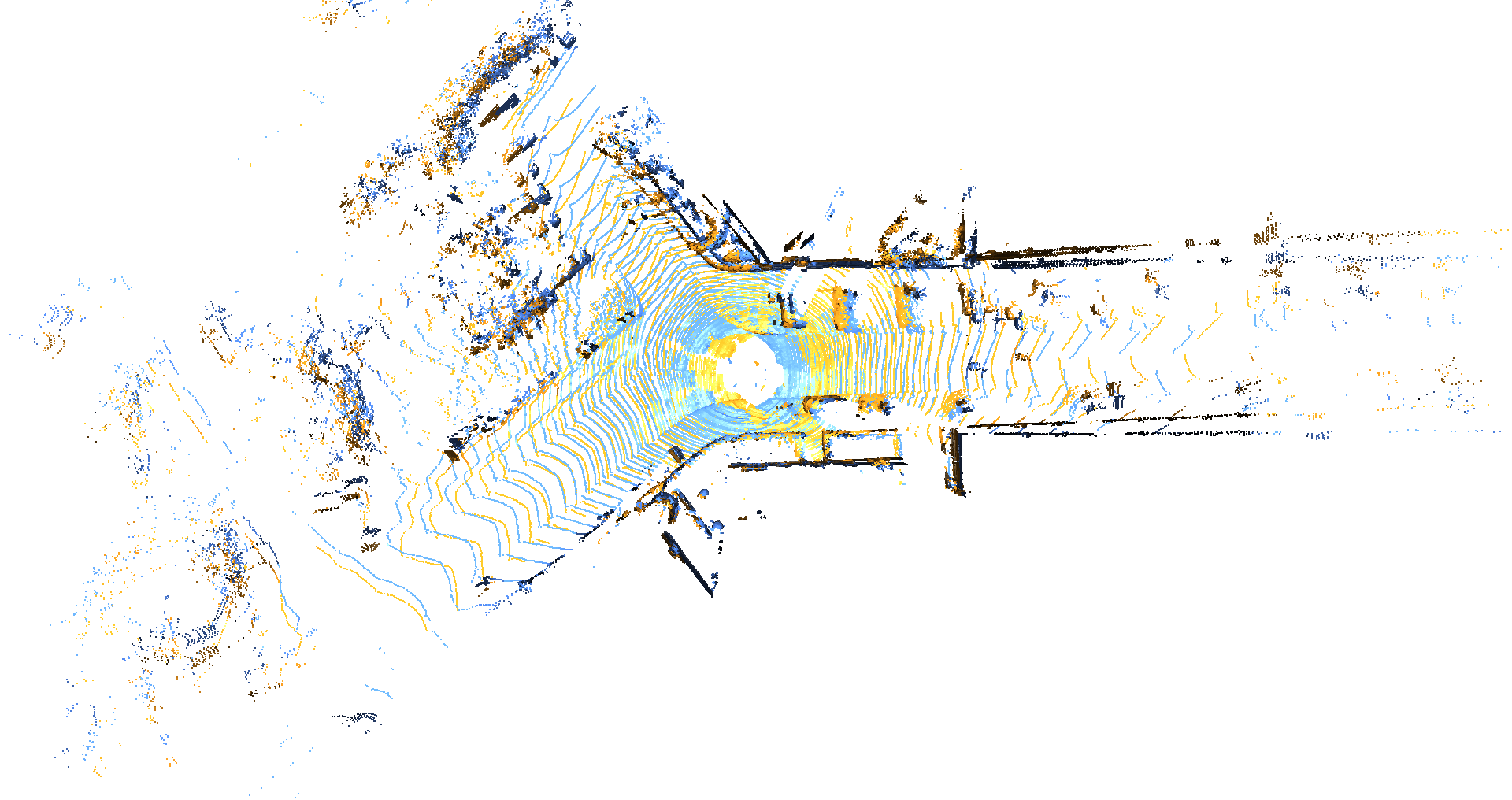} \\
    \ssf{(a) Point2Plane~\cite{Zhou2018}} & \ssf{(c) FGR~\cite{Zhou2016}} & \ssf{(e) Synthetic} \\
    \includegraphics[width=.32\columnwidth,trim={0 5cm 5cm 0},clip]{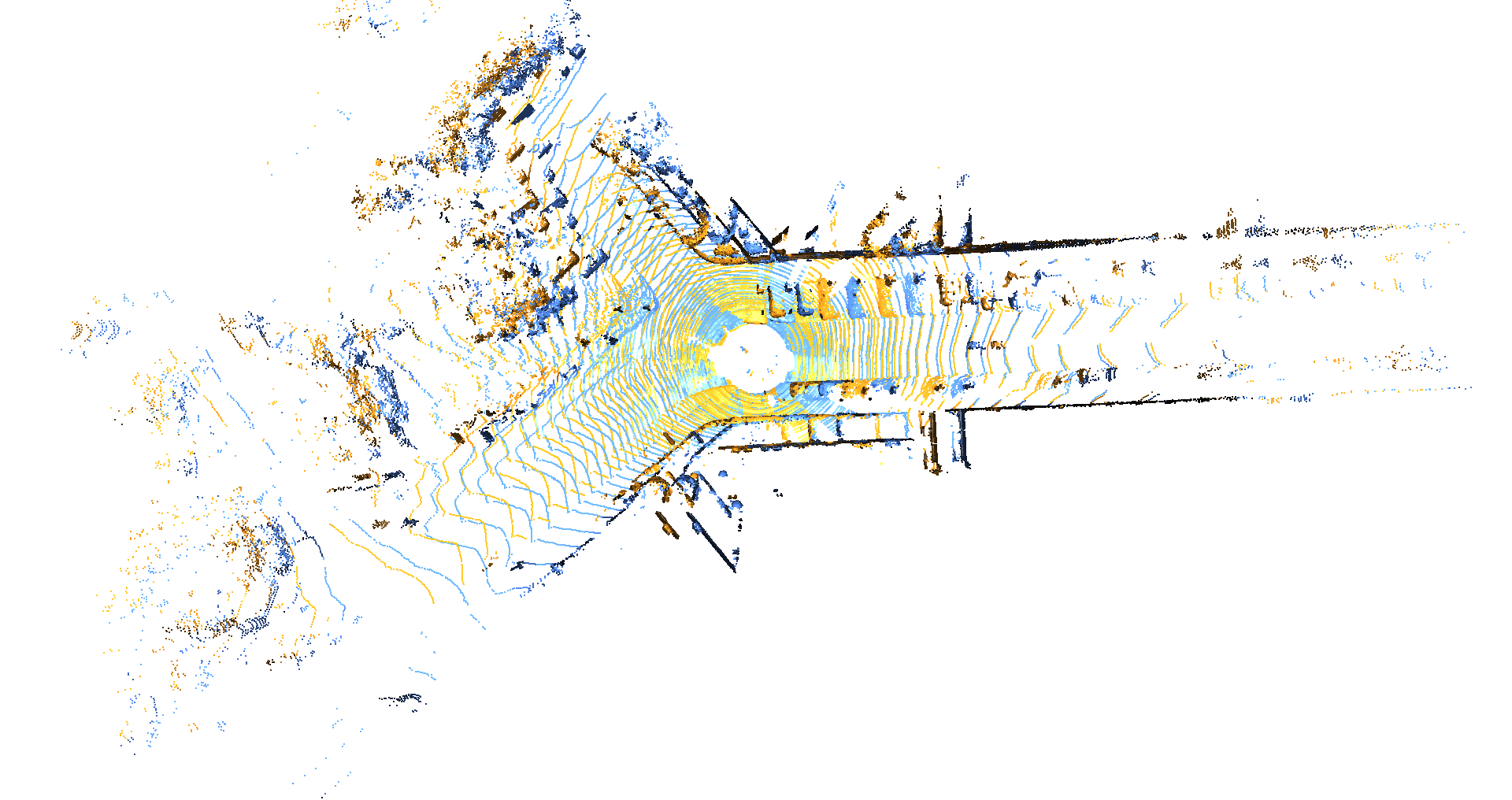} &
    \includegraphics[width=.32\columnwidth,trim={0 5cm 5cm 0},clip]{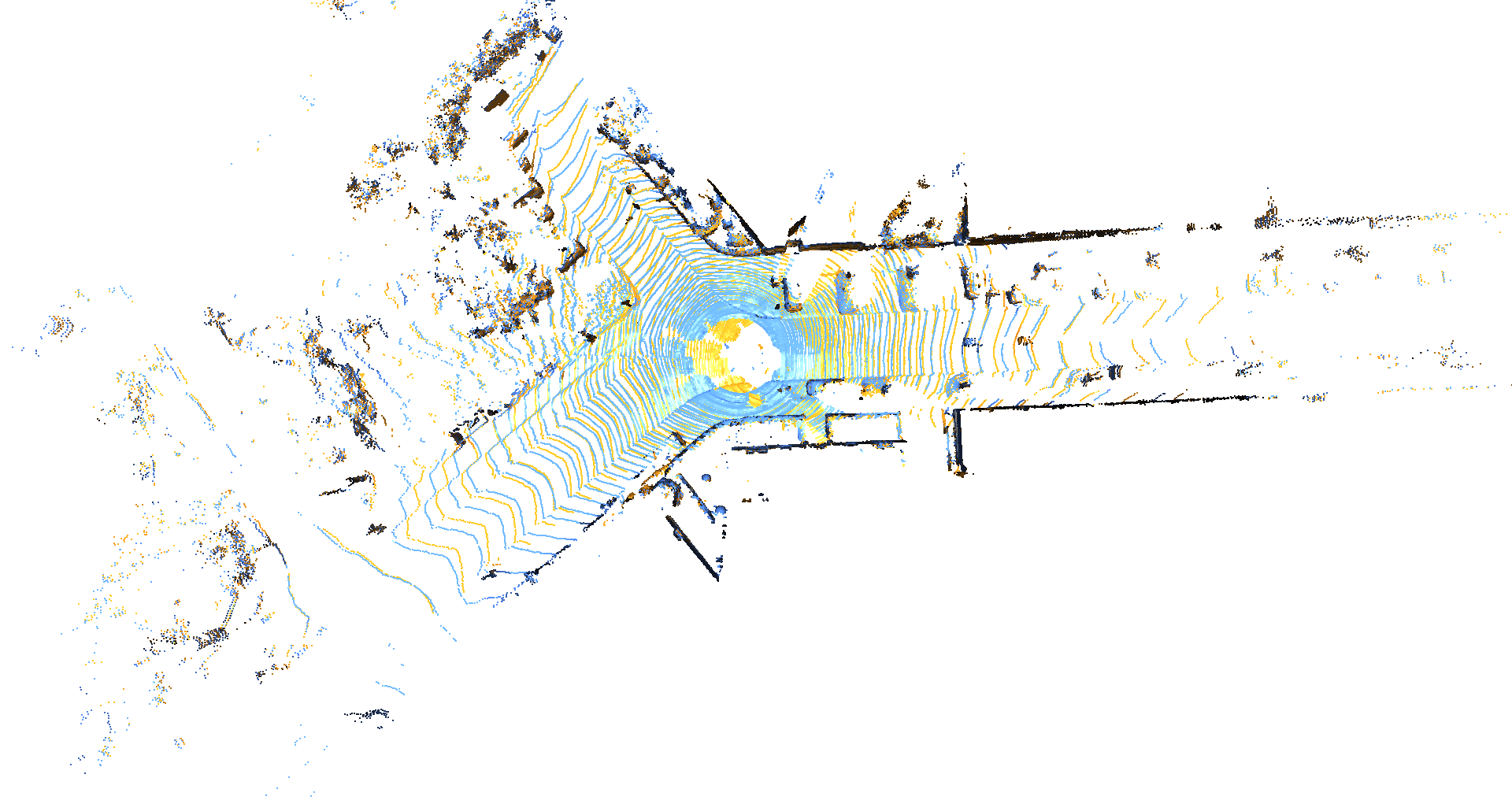} &
    \includegraphics[width=.32\columnwidth,trim={0 5cm 5cm 0},clip]{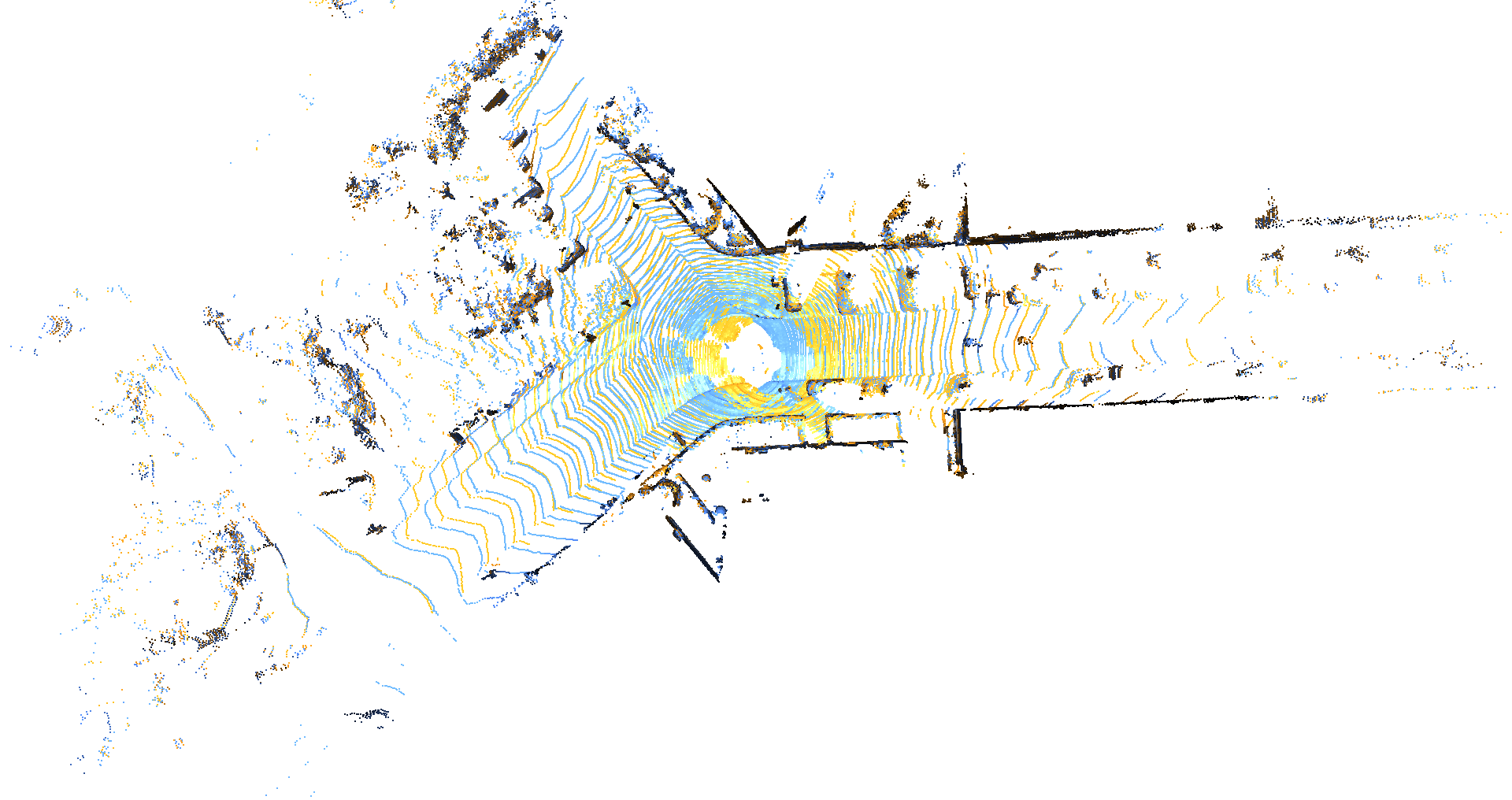} \\
    \ssf{(b) G-ICP~\cite{segal2009generalized}} & \ssf{(d) RANSAC~\cite{rusu2009fast}} & \ssf{(f) Synthetic-Multi}
    \end{tabular}
    \caption{Illustration of registration results on KITTI. While ICP variants converge to inaccurate transformations, projective registration with synthetic intrinsics results in better estimates, and can be further refined by multi-scale registration.}
    \vspace{-2mm}
    \label{fig:kitti-odometry}
\end{figure}

\begin{figure}[t]
    \centering
    \begin{tabular}{c}
    \includegraphics[width=.72\textwidth]{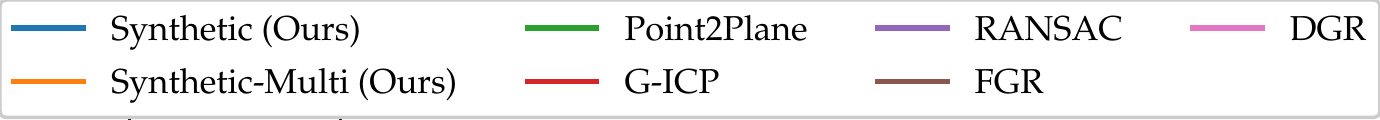}\\
    \includegraphics[width=.95\columnwidth, trim={.4cm 0 0 0}, clip]{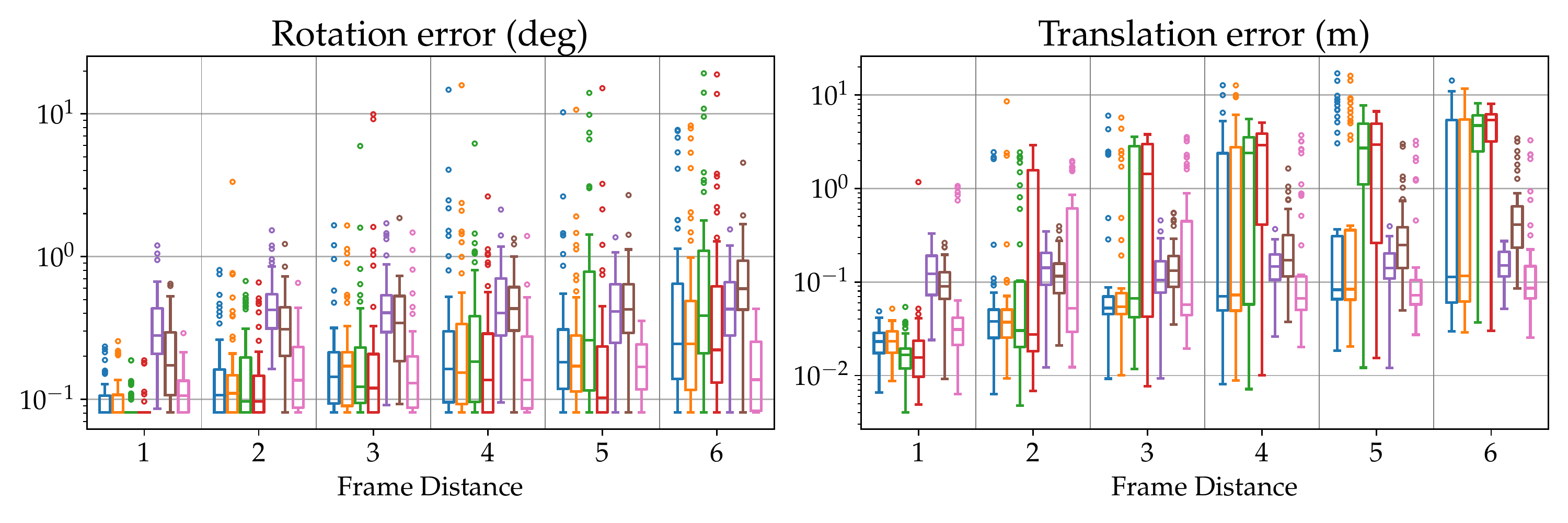}
    \end{tabular}
    \caption{Registration accuracy evaluation on KITTI sequence 00 with sampled pairs of enumerated frame differences. For each box plot, a lower median and smaller rectangle box is better. Without a LUT, projective registration with synthetic LiDAR intrinsics achieves comparable performance to ICP variants and global approaches.}
    \vspace{-3mm}
    \label{fig:exp-registration-kitti}
\end{figure}

\subsection{Baselines and Experimental Setups}
\subsubsection{Registration.}
We denote our approach with \emph{LUT} and \emph{LUT-Multi}, when intrinsic LUTs are available, and their simplified versions~\cite{Behley2018EfficientSS, shan2018lego} denoted by \emph{Synthetic} and \emph{Synthetic-Multi} with synthetic intrinsics. We select point-to-plane (\emph{Pt2Pl}) and \emph{G-ICP}~\cite{segal2009generalized} as ICP-variant baselines, and fast global registration (FGR)~\cite{Zhou2016}, RANSAC~\cite{rusu2009fast} as global registration baselines. We also compare against deep global registration (DGR)~\cite{choy2020deep} pretrained on KITTI, one of the state-of-the-art learning-based registration approaches.
In all experiments, we run 50 iterations for ICP variants and single-scale projective registration, \{20, 20, 10\} iterations for 3-level multi-scale registration, 1M iterations for RANSAC, and the default 64 iterations for FGR.
For a controlled comparison, we estimate normals in the point cloud form with radius nearest neighbor search, but organize them in the image domain.
An accelerated computation of normal map directly from range image~\cite{newcombe2011kinectfusion} is also available.
We conduct experiments on real-world sequences with enumerated frame distances, defined by $|j - i|$ for frame $i$ and $j$. A larger frame distance indicates a more challenging registration task. We use rotation error $e(\RR, \RR_{\mathrm{gt}}) = \arccos\frac{\RR \RR_{\mathrm{gt}}^\top - 1}{2}$ and translation error $e(\tt, \tt_{\mathrm{gt}}) = \lVert \tt - \tt_{\mathrm{gt}}\rVert^2$ as the evaluation metric. At each frame distance, we sample $M = 50$ pairs and compute the errors.
The distance threshold, serving as the radius for neighbor search in baselines, and the robust psuedo-Huber kernel size for our approaches, is 0.5m for outdoor scenes (KITTI and our dataset), and 0.2m for indoor scenes (our dataset).

\subsubsection{Surface reconstruction.}
We also compare our surface reconstruction module against volumetric reconstruction pipelines that supports LiDAR data, namely voxblox~\cite{oleynikova2017voxblox} (outputs triangle meshes) and Octomap~\cite{hornung2013octomap} (outputs point clouds). For evaluation, we use F-score computed by $F = \frac{\mathrm{precision} \cdot \mathrm{recall}}{\mathrm{precision}+\mathrm{recall}}$, where precision defines the percentage of points in the reconstruction with valid correspondences in the GT point cloud, and recall is the opposite. Unless mentioned, we use 0.1m as the voxel size for outdoor scenes, and 4cm for indoor scenes. We clip faraway points to maintain a reasonable memory footage and filter potential outliers. For indoor scenes and outdoor scenes the clipping distances are 10m and 30m, respectively.

\subsubsection{Implementation.} All the experiments are conducted on a machine with an NVIDIA RTX 3060 graphics card and an 16 core Intel i7-11700 CPU. The code is written in C++/CUDA with modularized python bindings.

\subsection{KITTI Dataset}
Before going through the evaluation on our collected dataset, we first briefly evaluate on the KITTI dataset~\cite{Geiger2012CVPR} with synthetic intrinsics to demonstrate the compatibility of our algorithms to point clouds. We deliver qualitative and quantitative registration experiments and provide qualitative reconstruction results. 
\vspace{-3mm}

\subsubsection{Registration.} 
Fig.~\ref{fig:kitti-odometry} shows the qualitative registration results. We observe that with challenging translation, the projective association ensures a wider search range for correspondences, and results in better convergence especially enhanced with multi-scale processing. We also quantitatively evaluate the registration accuracy with varying frame distances. Due to the fast moving speed, we limit the frame distance to 6 (otherwise overlaps between point clouds are limited). Here we use poses obtained from CMRNet~\cite{cattaneo2019cmrnet, chang2021map} as refined GT poses.

\begin{wrapfigure}{l}{.55\columnwidth}
    \centering
    \begin{tabular}{c}
        \includegraphics[width=0.55\textwidth, trim={0 3cm 0 0}, clip]{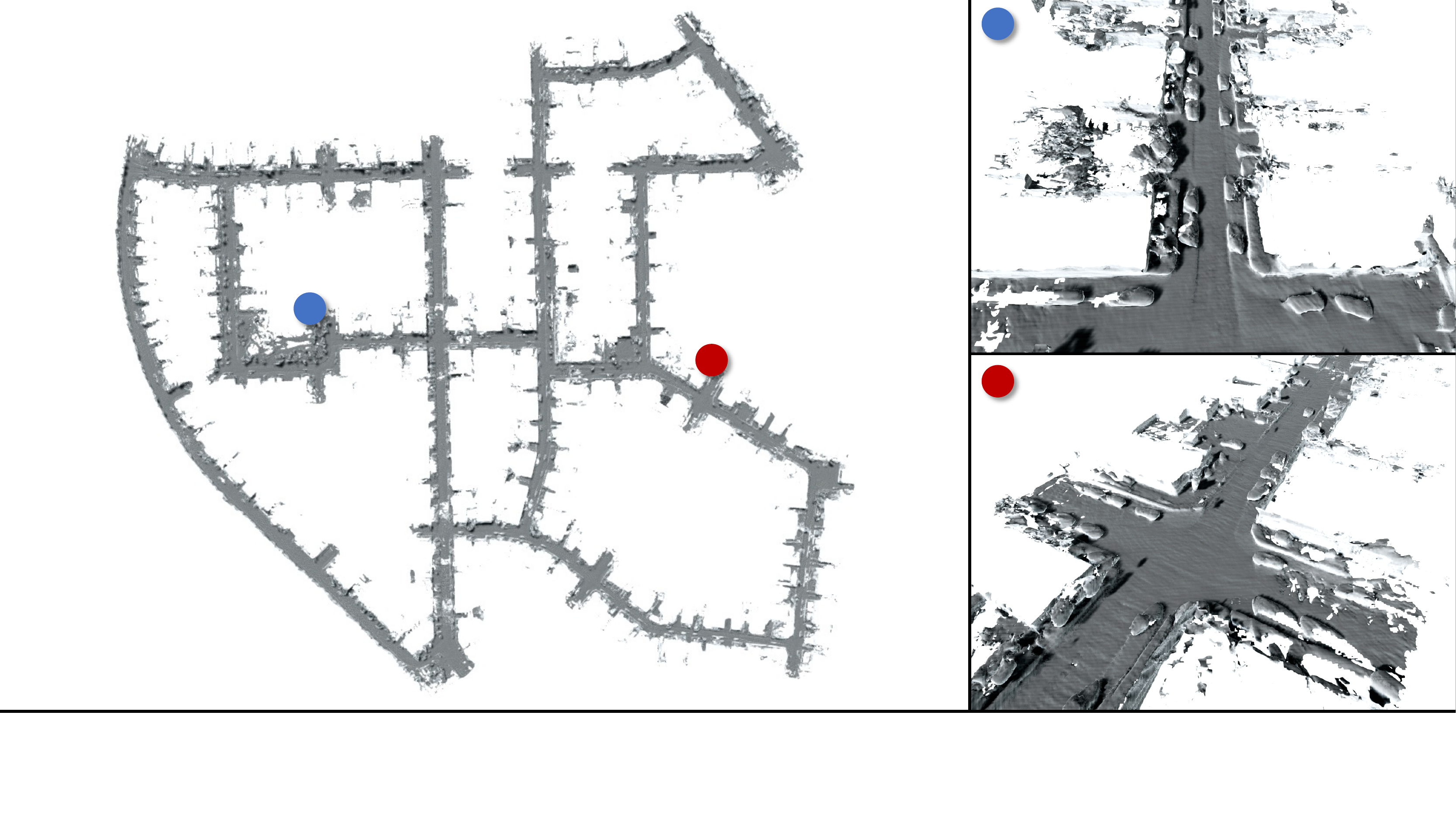} \\
        \includegraphics[width=0.55\textwidth, trim={0 3cm 0 0}, clip]{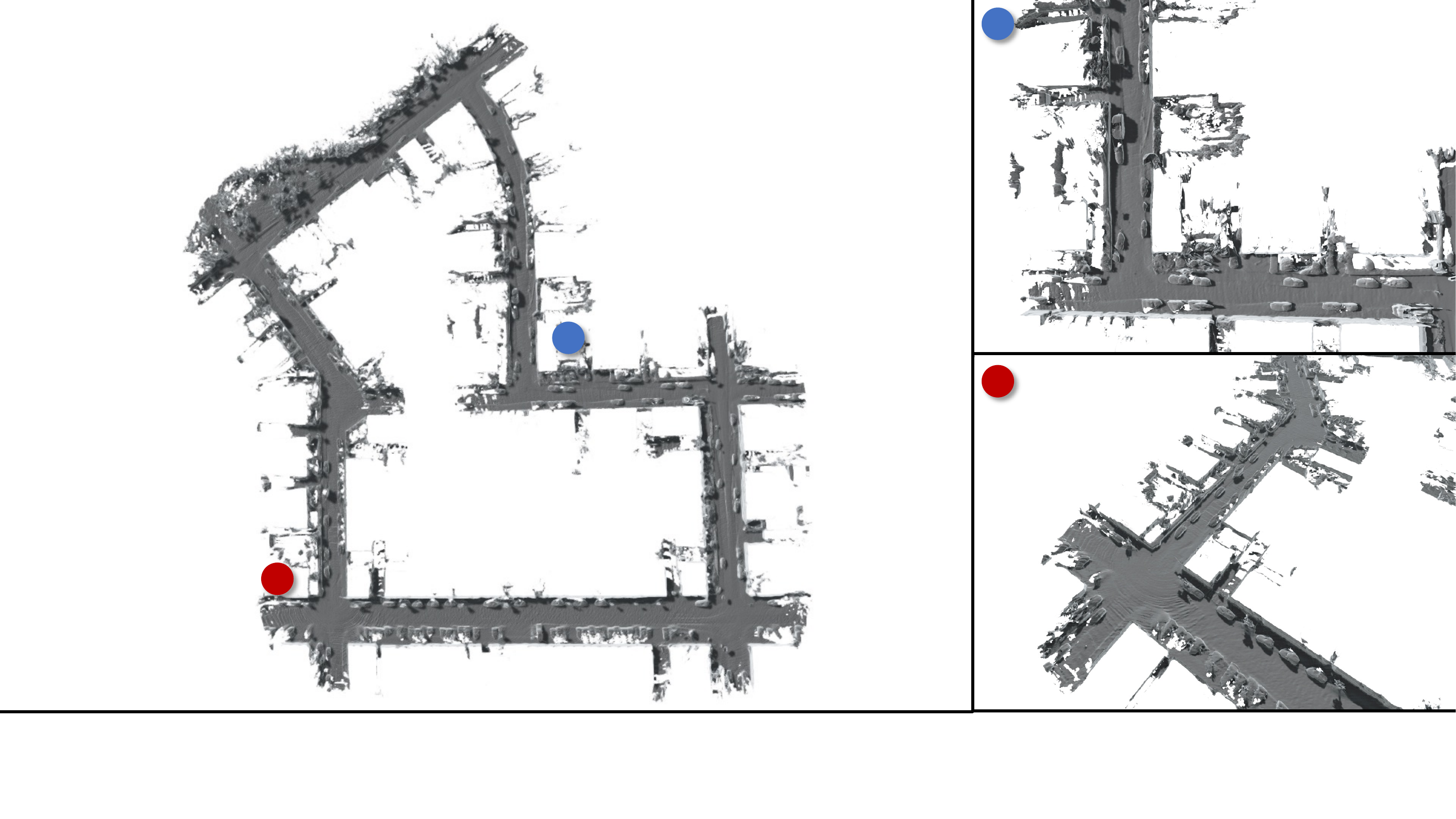}
    \end{tabular}
    \caption{City-scale TSDF surface reconstruction of sequences 00 and 07 from KITTI. Left: full reconstruction. Right: selected details. }
    \label{fig:kitti-reconstruction}
    \vspace*{-4mm}
\end{wrapfigure}
In Fig.~\ref{fig:exp-registration-kitti} we can observe that in comparison to ICP variants, projective registration has a comparable performance on translation and is consistently better on rotation due to the cylindrical presentation's advantage of a wide azimuth receptive field. At large frame distances, their medium rotation and translation error are also comparable to global registration results, including learning-based DGR.
\vspace{-3mm}
\subsubsection{Surface Reconstruction.}
In Fig.~\ref{fig:kitti-reconstruction} we show the meshes extracted from TSDF reconstruction at city scale on LiDAR data. With a limited GPU memory budget, we are able to reconstruct scenes with a 20cm voxel size at 40 Hz, where points farther than 30m are clipped.

%


\subsection{LiDAR Range Image Dataset}
\subsubsection{Registration.}
\begin{figure*}[t]
    \centering
    \begin{tabular}{cc}
    \includegraphics[width=.9\textwidth]{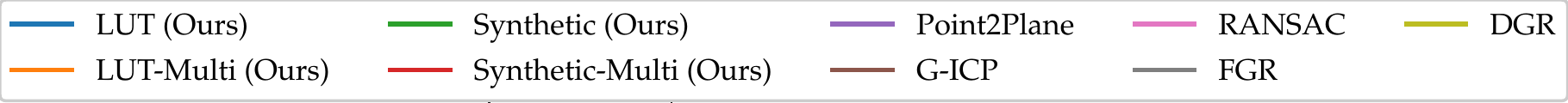}\\
    \includegraphics[width=.99\columnwidth, trim={.4cm 0 0 0}, clip]{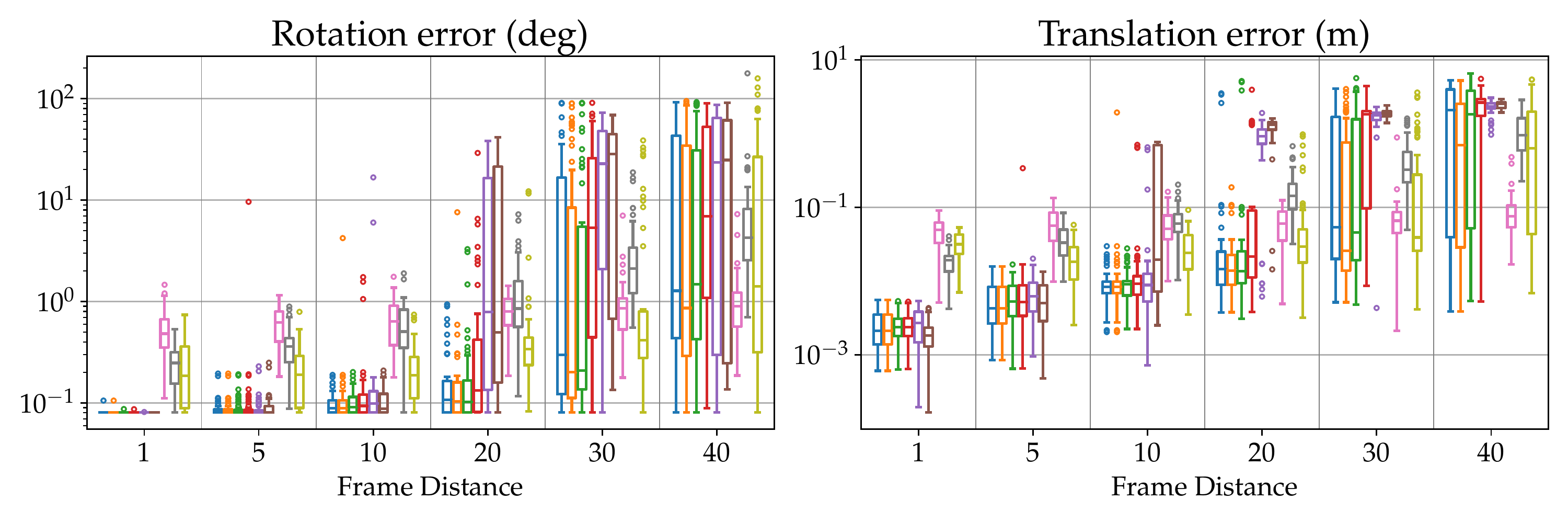} \\
    \includegraphics[width=.99\columnwidth, trim={.4cm 0 0 0}, clip]{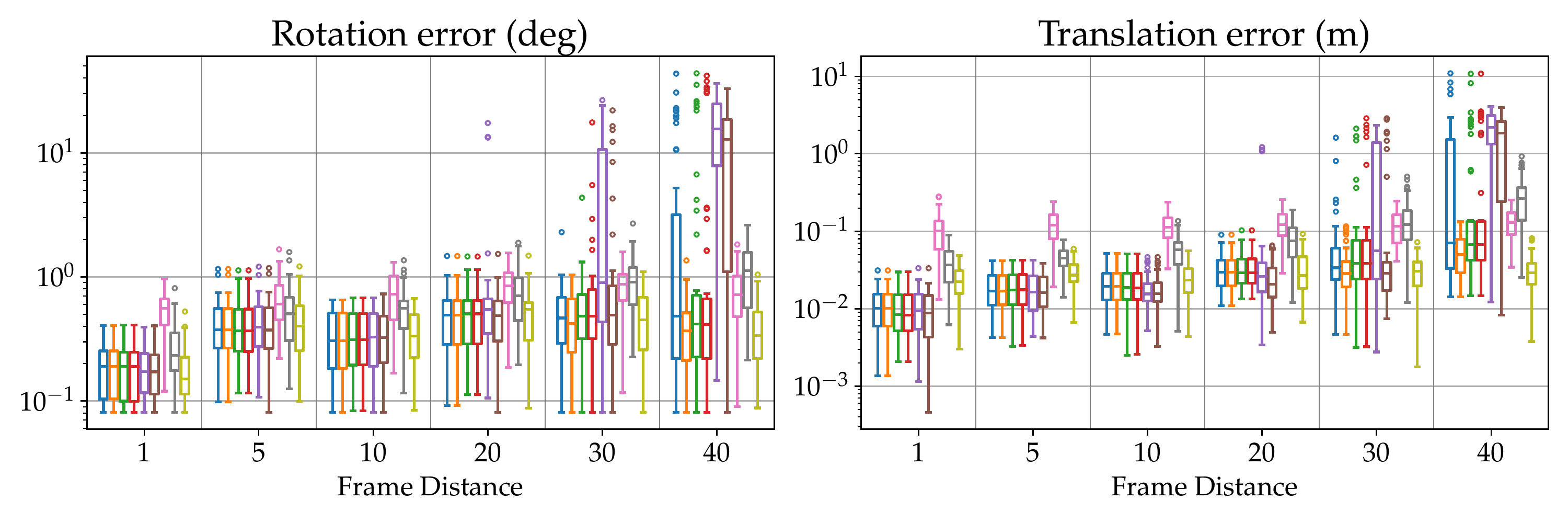} \\
    \end{tabular}
    \caption{Registration accuracy evaluation on the indoor \emph{lecture building} (top) and the outdoor \emph{dormitory} (bottom) sequences with sampled pairs of enumerated frame differences. For each box plot, a lower median and smaller rectangle box is better. Projective registration achieves comparable performance to baselines in general. Multi-scale projective registration equipped with an LUT is the best, with a stable performance at small frame distances and a better performance than even global approaches at large frame distances.}
    \vspace{-3mm}
  \label{fig:exp-registration-postech}
\end{figure*}

On our LiDAR range image dataset, we select a typical indoor scene~\emph{lecture building} and outdoor scene~\emph{dormitory}. The results are shown in Fig.~\ref{fig:exp-registration-postech}. In general, regardless of indoor or outdoor setups, we observe that range image based projective registration is comparable to ICP variants with small frame distances, and achieves better performance on more challenging registration tasks with large frame distances. At such setups, multi-scale registration with an LUT even outperforms global registration, including DGR, in most scenarios.

\begin{table}[ht]
\centering
\caption{F-score of surface reconstruction. Our method is consistently the best. }
\label{tab:reconstruction}
\resizebox{.9\textwidth}{!}{
\begin{tabular}{l| ccccc | ccccc}
\toprule
 & \rotentry{90}{\it\ssf{Lab}} & \rotentry{90}{\it\ssf{Lounge}} & \rotentry{90}{\it\ssf{Lecture Bld.}}
 & \rotentry{90}{\it\ssf{Lecture Rm.}} & \rotentry{90}{\it\ssf{Student Rm.}}
 & \rotentry{90}{\it\ssf{Campus}} & \rotentry{90}{\it\ssf{Fountain}} & \rotentry{90}{\it\ssf{Statue}}
 & \rotentry{90}{\it\ssf{Dormitory}} & \rotentry{90}{\it\ssf{Square}}\\
\midrule
\ssf{Voxblox~\cite{oleynikova2017voxblox}}
& \ssf{0.4845}   & \ssf{0.4919} & \ssf{0.4923}
& \ssf{0.4956}   & \ssf{0.4856}
& \ssf{0.4896}   & \ssf{0.4916} & \ssf{0.4885}
& \ssf{0.4901}   & \ssf{0.4893} \\
\ssf{Octomap~\cite{hornung2013octomap}}
& \ssf{0.4762}   & \ssf{0.4868} & \ssf{0.4526}
& \ssf{0.4948}   & \ssf{0.3883}
& \ssf{0.4090}   & \ssf{0.4588} & \ssf{0.4890}
& \ssf{0.4898}   & \ssf{0.4573} \\
\midrule
\ssf{Ours}
& \ssbf{0.4900}  & \ssbf{0.4940}  & \ssbf{0.4930}
& \ssbf{0.4972}  & \ssbf{0.4870}
& \ssbf{0.4907}  & \ssbf{0.4933}  & \ssbf{0.4895}
& \ssbf{0.4917}  & \ssbf{0.4901}\\
\bottomrule
\end{tabular}
}
\end{table}

\subsubsection{Surface Reconstruction.}
%

In Fig.~\ref{fig:dataset}, we qualitatively show the reconstructed surfaces from SDF volumes overlaid with the camera trajectory. We observe that in indoor scenes, our algorithm reconstructs high quality surfaces, despite our range images having lower spatial density. In addition, we are able to reconstruct high quality surfaces of large scale outdoor scenes.

Quantitative results are shown in Table~\ref{tab:reconstruction}, where the valid correspondence searching range is set to 3$\times$ voxel size for precision and recall computation in F-score. Comparing to the baselines, our reconstruction achieves consistently the highest F-score. 



\subsubsection{Runtime Evaluation.}
We then demonstrate the efficiency of our approaches by evaluating average run time on indoor and outdoor scenes separately. In Table~\ref{tab:runtime} we observe that while achieving comparable or better accuracy than state-of-the-art approaches, our method is 15--50$\times$ faster in volumetric reconstruction, and 5--150$\times$ faster in registration.
\begin{table}[t]
\centering
\caption{Run time evaluation. With parallel projective operations implemented on GPU, our methods are at least one magnitude faster than baselines.}\label{tab:runtime}
\resizebox{.9\textwidth}{!}{
\begin{tabular}{l| ccc | ccccccc}
\toprule
 & \rotentry{90}{\emph{\ssf{Voxblox}}~\ssf{\cite{oleynikova2017voxblox}}} & \rotentry{90}{\emph{\ssf{Octomap}}~\ssf{\cite{hornung2013octomap}}} & \rotentry{90}{\it\ssf{Ours}} 
 & \rotentry{90}{\emph{\ssf{Pt2Pl}}~\ssf{\cite{Zhou2018}}} & \rotentry{90}{\emph{\ssf{G-ICP}}~\ssf{\cite{segal2009generalized}}}
 & \rotentry{90}{\emph{\ssf{RANSAC}}~\ssf{\cite{rusu2009fast}}} & \rotentry{90}{\emph{\ssf{FGR}}~\ssf{\cite{Zhou2016}}} & \rotentry{90}{\emph{\ssf{DGR}}~\ssf{\cite{choy2020deep}}}
 & \rotentry{90}{\it\ssf{Ours (LUT)}} & \rotentry{90}{\it\ssf{Ours (LUT-multi)}}\\
 \midrule
 \ssf{Indoor} &  \ssf{1003.26} & \ssf{676.20} & \ssbf{22.29}
 & \ssf{62.04} & \ssf{118.61}
 & \ssf{445.97} & \ssf{977.56} & \ssf{347.39}
 & \ssf{14.14} & \ssbf{10.90}\\
 \ssf{Outdoor}  & \ssf{1002.22} & \ssf{1167.68} & \ssbf{58.55}
 & \ssf{59.19} & \ssf{118.60}
 & \ssf{408.16} & \ssf{1407.59} & \ssf{1184.51}
 & \ssf{14.20} & \ssbf{10.69}\\
\bottomrule
\end{tabular}
}
\end{table}

\subsection{Limitations}
Our registration method has presented benefits both in efficiency and accuracy in the experiments, yet there are several limitations. Although it has achieved good performance with challenging rotations, it still has a reduced stability on large translations that cannot be addressed by the cylindrical representation. Since fundamentally it is depending on dense nonlinear optimization, similar to other local registration approaches, it may also fall into local optima when the scene's structures are not salient. 
Another limitation is that the projective model can be disturbed by fast sensor motions and dynamic environments, where the projection model needs modification for moving ray centers.
 In the future, we would attempt to learn deep cylindrical image features for global feature matching using \eg ~Spherical CNNs~\cite{cohen2018spherical}; we will also consider non-rigid transform with consecutive poses assigned to each column~\cite{zhang2014loam}. 
 

\section{Conclusion}
\label{sec:conclusion}
We presented a range image based LiDAR data representation from raw sensor data that naturally preserves the neighbor information. With an intrinsic spherical projective model, it allows us to perform fast and accurate range image based multi-scale registration and dense reconstruction. We then collected a new LiDAR dataset in the image form, and perform comprehensive experiments for dense reconstruction and registration demonstrating the efficiency of our approaches.
With the proof of concept, we humbly hope the hardware manufacturers may expose more user-friendly interfaces to generate LiDAR images for fast and accurate 3D perception, and the vision community may find it easier to transfer the knowledge from 2D to 3D.

\clearpage
{\small
	\bibliographystyle{splncs04}
	\bibliography{paper}
}

\end{document}